\documentclass{article}
\usepackage{amsthm}
\usepackage{thm-restate}
\usepackage{booktabs}
\usepackage{framed}
\usepackage{color}
\usepackage{xcolor}
\definecolor{mydarkblue}{rgb}{0,0.08,0.45}
\usepackage[hidelinks,colorlinks]{hyperref}
\hypersetup{
	colorlinks=true,
	linkcolor=mydarkblue,
	citecolor=mydarkblue,
	filecolor=mydarkblue,
	urlcolor=mydarkblue,
}
\usepackage[ruled]{algorithm2e}
\usepackage{thmtools,thm-restate}
\theoremstyle{definition}

\newtheorem*{theorem*}{Theorem}

\usepackage{times}

\usepackage{enumerate}

\usepackage{graphicx}
\usepackage[numbers]{natbib}
\usepackage{mathrsfs,amssymb,amsmath,textcomp,amsfonts,bbm}
\usepackage{color}
\usepackage{multirow,amsmath, dsfont, amscd, latexsym,color,amssymb,setspace}
\usepackage{ifthen}
\usepackage{enumerate}

\usepackage{listings}
\usepackage{enumitem}
\usepackage{etoolbox}
\usepackage{caption}
\usepackage{thmtools,thm-restate}
\usepackage{comment}
\usepackage{subcaption}
\usepackage{url}
\usepackage{graphicx,caption, subcaption, multirow}
\usepackage{float}
\providetoggle{solution}
\settoggle{solution}{false}

\def\P{\mathcal{P}}

\def\T{\mathcal{T}}

\newcommand{\TODO}[1]{}

\newcommand{\remove}[1]{}

\newcommand{\defeq}{:=}

\newtheorem{Thm}{Theorem}
\newtheorem{Lem}[Thm]{Lemma}

\newtheorem{Def}[Thm]{Definition}

\newtheorem*{Conj*}{Conjecture}

\newenvironment{Proof}{\medbreak
\noindent {\bf Proof:~}}{\unskip\nobreak\hfill\hskip 2em \qed\par\medbreak}

\newcommand{\eat}[1]{}

\newcommand{\ol}[1]{\ensuremath{\bar{#1}}}

\newcommand{\rgta}{\ensuremath{\rightarrow}}

\newcommand{\zo}{\{0,1\}}

\newcommand{\intzo}{[0,1]}

\newcommand{\commented}{no}

\ifthenelse{\equal{\commented}{yes}}{
\newcommand{\pnote}[1]{\footnote{{\bf [[Parik: {#1}\bf ]] }}}
}

\ifthenelse{\equal{\commented}{no}}{
\newcommand{\pnote}[1]{}
}

\newcommand{\ignore}[1]{}

\def\colorful{3}

\ifnum\colorful=1

\fi

\newcommand{\Var}{{\bf Var}}

\newcommand{\id}{\mathrm{idLength}}
\newcommand{\ID}{\mathrm{ID}}
\newcommand{\pid}{\mathrm{pidLength}}
\newcommand{\PID}{\mathrm{PID}}
\newcommand{\pidsc}{\mathrm{PIDScore}}
\newcommand{\Imp}{\mathrm{Imp}}
\newcommand{\vol}{\mathrm{vol}}
\newcommand{\len}{\mathrm{len}}
\newcommand{\cost}{\mathrm{cost}}
\newcommand{\iF}{\rm iForest}
\newcommand{\pidF}{\rm PIDForest}

\newcommand{\I}{{\mathbb{I}}}
\newcommand{\R}{\mathbb R}

\newcommand{\eps}{\varepsilon}

\newcommand{\myalgo}[2]{
\bigskip
\noindent \fbox{
\begin{minipage}{5.3in}
{\bf #1}
{\tt #2}
\end{minipage}}
\bigskip
}

\newcounter{this-list}

\newif\ifdraft
\draftfalse
\ifdraft
\newcommand{\udi}[1]{{\color{blue}\textbf{Udi: #1}}}
\newcommand{\parikshit}[1]{{\color{red}\textbf{Parikshit: #1}}}
\newcommand{\vatsal}[1]{{\color{green}\textbf{Vatsal: #1}}}
\newcommand{\VS}[1]{\textcolor{green}{VS: #1}}
\newcommand{\PG}[1]{\textcolor{blue}{PG: #1}}
\newcommand{\UW}[1]{\textcolor{brown}{UW: #1}}
\newcommand{\Rev}[1]{\textcolor{red}{Reader 1 : #1}}
\newcommand{\Revtwo}[1]{\textcolor{magenta}{Reader 2 : #1}}
	\else
	\newcommand{\udi}[1]{}
	\newcommand{\parikshit}[1]{}
	\newcommand{\vatsal}[1]{}
	\newcommand{\VS}[1]{}
	\newcommand{\PG}[1]{}
	\newcommand{\UW}[1]{}
	\newcommand{\Rev}[1]{}
	\newcommand{\Revtwo}[1]{}
\fi
\newcommand{\density}{PIDForest }
     \usepackage[final]{neurips_2019}

\usepackage[utf8]{inputenc} % allow utf-8 input
\usepackage[T1]{fontenc}    % use 8-bit T1 fonts
\usepackage{hyperref}       % hyperlinks
\usepackage{url}            % simple URL typesetting
\usepackage{booktabs}       % professional-quality tables
\usepackage{amsfonts}       % blackboard math symbols
\usepackage{nicefrac}       % compact symbols for 1/2, etc.
\usepackage{microtype}      % microtypography

\title{PIDForest: Anomaly Detection via Partial Identification}

\author{
  Parikshit Gopalan\\
  VMware Research\\
  \texttt{pgopalan@vmware.com} \\
  \And
  Vatsal Sharan\thanks{Vatsal's contribution were supported by NSF award 1813049.} \\
  Stanford University\\
  \texttt{vsharan@stanford.edu}
  \And
  Udi Wieder\\
  VMware Research\\
  \texttt{uwieder@vmware.com}
}

\begin{document}

\maketitle

\begin{abstract}
We consider the problem of detecting anomalies in a large dataset. We
propose a framework called Partial Identification which captures the
intuition that anomalies are easy to distinguish from the overwhelming
majority of points by relatively few attribute values. Formalizing
this intuition, we propose a geometric anomaly measure for a point that we call $\pidsc$,
which measures the minimum density of data points over all subcubes
containing the point. We present \pidF: a random forest based
algorithm that finds anomalies based on this definition. We show that
it performs favorably in comparison to several popular anomaly
detection methods, across a broad range of benchmarks. \pidF\ also
provides a succinct explanation for why a point is labelled anomalous,
by providing a set of features and ranges for them which are
relatively uncommon in the dataset.

\end{abstract}
\section{Introduction}

An anomaly in a dataset is a point that does not conform to what is normal or
expected.  Anomaly detection is a ubiquitous machine learning task
with diverse applications including network monitoring, medicine and
finance \cite[Section 3]{ChandolaBK09}. There is an extensive body of research devoted to it, see
\cite{ChandolaBK09, Aggarwal2013} and the references therein. Our work is primarily motivated by the emergence of large distributed
systems, like the modern data center which produce massive amounts of
heterogeneous data. Operators need to constantly monitor this data, and use it to identify and troubleshoot problems. The volumes of data involved are so large that a lot of the analysis has to be automated. \eat{Effective anomaly detection algorithms can help humans prioritize attention and
hone in on important events.} Here we highlight some of the challenges
that an anomaly detection algorithm must face.

\begin{enumerate}
\item 	{\bf High dimensional, heterogeneous data:}
          The data collected could contains measurements of metrics like cpu usage, memory, bandwidth, temperature,
          in addition to categorical data such as day of the
          week, geographic location, OS type. This makes finding an
          accurate generative model for the data challenging. The metrics might be
          captured in different units, hence algorithms that are
          unit-agnostic are preferable. The algorithm needs to scale
          to high dimensional data.
          %\Rev{Language could be a bit more formal, not using 'think',
          %'etc', 'like'}

	\item {\bf Scarce labels: } Most of the
          data are unlabeled. Generating labels is time
          and effort intensive and requires domain knowledge. Hence supervised methods are a
          non-starter, and even tuning too many hyper-parameters of
          unsupervised algorithms could be challenging.

        \item {\bf Irrelevant attributes:} Often an anomaly manifests
          itself in  a relatively small number of
          attributes among the large number being monitored. For
          instance, a single machine in a large datacenter might be
          compromised and behave abnormally.

         \item {\bf Interpretability of results: } When we alert a
          datacenter administrator to a potential anomaly, it helps
          to point to a few metrics that might have triggered it, to help in troubleshooting.
%	\item Even though the data is often in the form of a time-series, many of the most interesting anomalies are stationary in nature.
\end{enumerate}

In the generative model setting, anomalies come with a simple explanation: a
model that fits the data, under which the anomalous
observation is unlikely. Interpretability is more challenging
for algorithms that do not assume a generative model. In this work, we are
particularly interested in random forest based methods for anomaly
detection, namely the influential work on Isolation Forests
\cite{iForest} (we refer to this algorithm as \iF) and subsequent work
\cite{GuhaMRS16, BandaragodaTALZ18}. \iF\ is a remarkably simple and efficient algorithm,
that has been found to outperform other anomaly
detection methods in several domains \cite{emmott2013systematic}.
Yet, there is no crisp definition of ground truth for what constitutes
an anomaly: the anomaly score is more or less {\em defined} as the
output of the algorithm.  We believe that a necessary step for
interpretability is a clear articulation of {\em   what} is an
anomaly, separate from the algorithmic question of {\em how} it is found.

\paragraph{Our contributions.} We summarize the main contributions of this work:
\begin{enumerate}
  \item In Section \ref{sec:pidsc}, we  motivate and propose a  new anomaly measure that we call
    $\pidsc$. Our definition corresponds to an intuitive notion of
    what is an anomaly and has a natural geometric interpretation. It
    is inspired by the notion of Partial Identification introduced by \citet{WigdersonY16}  can be viewed as a
    natural generalization of
    teaching dimension \cite{GoldmanK95,  KushilevitzLRS96}.

\item Our definition sheds light on
the types of points likely to be labeled as anomalies by the
\iF\ algorithm, and also on the types of points it might miss. We
build on this intuition to design an efficient random forest based algorithm---\pidF,
which finds anomalies according to $\pidsc$, in Section \ref{sec:pidf}.

  \item We present extensive experiments on real and synthetic data
    sets showing that our algorithm consistently outperforms or
    matches six popular anomaly detection algorithms. PIDForest is the
    top performing algorithm in $6$ out of $12$ benchmark real-world
    datasets, while no other algorithm is the best in more than
    $3$. \pidF\ is also resilient to noise and irrelevant attributes.
    These results are in Section \ref{sec:real} and \ref{sec:synthetic}.
\end{enumerate}

We begin by describing our proposed anomaly measure, $\pidsc$ at a
high level. Let the {\em sparsity} of a dataset $\T$ in a subcube of the attribute space be the
volume of the subcube divided by the number of points from $\T$ that it contains.
For a dataset $\T$ and a point $x$, $\pidsc(x, \T)$
measures the maximum sparsity of $\T$ in all subcubes $C$ containing $x$. A point $x$ is labelled
anomalous if it belongs to a region of the attribute space where data
points are sparse.

Given this definition, one could aim for an algorithm that
preprocesses $\T$, then takes a point $x$ and computes $\pidsc(x, \T)$.
Such an algorithm is likely to suffer from the curse of dimensionality
like in Nearest Neighbor based methods, and not scale to high volumes
of data.
Instead we adopt the approach of \iF\ \cite{iForest} which focuses on
what is anomalous, rather than the entire dataset. We call the
resulting algorithm \pidF.
Like in \iF, \pidF\ builds a collection of decision trees that
partition space into subcubes. In \pidF, the choice of the splits at
each node favors partitions of greatly varying sparsity, the variance
in the sparsity is explicitly the quantity we optimize when choosing a
split. In contrast, previous work either choose splits randomly
\cite{iForest} or based on the range \cite{GuhaMRS16}. Choosing
coordinates that have greater variance in their marginal distribution
lets us hone in on the important coordinates, and makes our algorithm robust to
irrelevant/noisy attributes, which are unlikely to be chosen. Secondly, we
label each leaf by its sparsity rather than depth in the tree. The
score of a point is the maximum sparsity over all leaves reached in
the forest.

While notions of density have been used in previous works on clustering
and anomaly detection, our approach differs from prior work in
important ways.

\begin{enumerate}
  \item {\bf Dealing with heterogeneous attributes: }
    Dealing with subcubes and volumes allows us to handle
    heterogeneous data where some columns are real, some are
    categorical and possibly unordered. All we need is to specify two
    things for each coordinate: what
    constitutes an interval, and how length is measured. Subcubes and
    volumes are then defined as products over coordinates.
    This is in sharp contrast to methods that assume a
    metric. Notions like $\ell_1/\ell_2$ distance
    add different coordinates and might not be natural in heterogeneous
    settings.

  \item {\bf Scale invariance:} For a subcube, we only care about the ratio
    of its volume to the volume of the entire attribute space. Hence
    we are not sensitive to the units of measurement.

  \item {\bf Considering subcubes at all scales:} In previous works,
    density is computed using balls of a fixed radius, this
    radius is typically a hyperparameter. This makes
   the algorithm susceptible to masking, since there may be a dense cluster
   of points, all of which are anomalous. We take the minimum over
   subcubes at all scales.
\end{enumerate}

\paragraph{Organization.} The rest of the paper is organized as follows. We present the
definition of $\pidsc$ in Section \ref{sec:pidsc}, and the
\pidF\ algorithm (with a detailed comparison to iForest) in Section \ref{sec:pidf}. % and a detailed comparison between \pidF\ and \iF\ in Section \ref{sec:if}.
We present experiments on real world
data in Section \ref{sec:real} and
synthetic data in Section \ref{sec:synthetic}. We further discuss related work in Section \ref{sec:related}.

%%%%%%%%%%%%%%%%%%%%%%%%%%%%%%%%%%%%%%%%%%%%%%%%%%%%%%%%%%%%%%%%%%%%%%%%%%%%%%%%%%%%%%%%%
%%%%     Boolean Setting
%%%%%%%%%%%%%%%%%%%%%%%%%%%%%%%%%%%%%%%%%%%%%%%%%%%%%%%%%%%%%%%%%%%%%%%%%%%%%%%%%%%%%%%%%

\section{Partial Identification and $\pidsc$}
\label{sec:pidsc}

{\bf A motivating example: Anomalous Animals.}
Imagine a tabular data set that contains a row for every animal on the
planet. Each row then contains attribute information about the animal
such as the species, color, weight, age and so forth. The rows are ordered.
Say that Alice wishes to identify a particular animal in the table
unambiguously to Bob, using the fewest number of bits.

If the animal happens to be a {\em white elephant}, then Alice is in
luck. Just specifying the species and color narrows the list
of candidates down to about fifty (as per Wikipedia). At this point, specifying
one more attribute like weight or age will probably pin the animal
down uniquely. Or she can just specify its order in the list.

If the animal in question happens to be a {\em white rabbit}, then
it might be far harder to uniquely identify, since there are tens of millions of
white rabbits, unless that animal happens to have some
other distinguishing features. Since weight and age are numeric rather
than categorical attributes, if one could measure them to arbitrary
precision, one might be able to uniquely identify each
specimen. However, the higher the precision, the more bits Alice needs
to communicate to specify the animal.

%\Rev{This precision issue for continuous data does not motivate it well.}

We will postulate  a formal definition of
anomaly score, drawing on the following intuitions:

\begin{enumerate}

  \item {\bf Anomalies have short descriptions.}
The more exotic/anomalous the animal Alice has in mind, the more it
stands out from the crowd and the easier it is for her to convey it to Bob.
    Constraining just a {\em few
      carefully chosen} attributes sets anomalies apart from the vast
    majority of the population.

  \item {\bf Precision matters in real values.}
    For real-valued attributes, it makes sense to specify a range in
    which the value lies. For anomalous points, this range might not
    need to be very narrow, but for normal points, we might need more
    precision.

  \item {\bf Isolation may be overkill.}
    The selected attributes need not suffice for complete
    isolation. {\em Partial identification} aka narrowing the space down to
    a small list can be a good indicator of an anomaly.

\end{enumerate}

First some notation: let $\T$ denote a dataset of $n$ points in $d$
dimensions. Given indices $S \subseteq [d]$ and $x \in \R^d$, let
$x_S$ denote the projection of $x$ onto coordinates in $S$. Logarithms
are to base $2$.

%\Revtwo{We should emphasize and make it clear that this is our contribution to propose these definitions. }

\subsection{The Boolean setting}
\label{sec:boolean}

We first consider the Boolean setting where the set of points is $\T \subseteq \zo^d$.
Assume that $\T$ has no duplicates.
We define $\id(x, \T)$ to be the minimum number of co-ordinates that
must be revealed to uniquely identify $x$ among all points in $\T$.
Since there are no duplicates, revealing all coordinates suffices, so
$\id(x, \T) \leq d$,

\begin{Def} (IDs for a point)
\label{def:boolean-id}
  We say that $S \subseteq [d]$ is an ID for $x \in \T$ if $x_S \neq
  y_S$ for all $y \in \T\setminus \{x\}$. Let $\ID(x, \T)$ be
  the smallest ID for $x$ breaking ties arbitrarily. Let $\id(x, \T)
  = |\ID(x, \T)|$.
\end{Def}

While on first thought $\id$ is an appealing measure of anomaly, it
does not deal with duplicates, and further, the
requirement of unique identification is fairly stringent. Even in
simple settings points might not have short IDs. For example, if $\mathcal{H}$
is the Hamming ball consisting of $0^d$ and all $d$ unit vectors, then $\id(0^d,
\mathcal{H}) = d$, since we need to reveal all the coordinates to separate
$0^d$ from every  unit vector. One can construct examples where even
the average value of $\id(x, \T)$ over all points can be surprisingly
high \cite{KushilevitzLRS96}.

We relax the definition to allow for \emph{partial identification}.
Given $x \in \T$ and $S \subseteq[d]$, the set of impostors of $x$ in
$\T$ are all points that equal $x$ on all coordinates in $S$. Formally
$\Imp(x, \T, S) = \{ y \in \T  \ s.t. \  x_S = y_S\}$.
We penalize sets that do not identify $x$ uniquely by the
{\em logarithm of
the number of impostors}. The intuition is that this penalty measures how many bits it
costs Alice to specify $x$ from the list of impostors.

\begin{Def}(Partial ID)
\label{def:boolean-pid}
    We define
  \begin{align}
    \label{eq:zo-def1}
    \PID(x, \T) & = \arg\min_{S \subseteq [d]} (|S| + \log_2(|\Imp(x, \T,
    S)|)), \\
  \label{eq:zo-def2}
  \pid(x, \T) & = \min_{S \subseteq [d]} (|S| + \log_2(|\Imp(x, \T,
  S)|)).
  \end{align}
\end{Def}

It follows from the definition that $\pid(x, \T) \leq \min(\log_2(n),
\id(x, \T))$.
The first inequality follows by taking $S$ to be empty so that every point
in $\T$ is an impostor, the second by taking $S = \ID(x, \T)$ so that
the only impostor is $x$ itself. Returning to the Hamming ball
example, it follows that $\pid(0^d, \T) = \log_2(d +1)$ where we take the
empty set as the $\PID$.

\eat{
{\bf Information Theoretic interpretation: } Let us return to Alice
and Bob trying to identify a unique animal in their table. Assume now
that the rows of their table are ordered.
To identify a particular
animal, Alice can specify a set of attributes, and then the index of the animal in the resulting
ordered list, which costs $\log_2(|\Imp|)$ bits. Thus $\pid(x, \T)$ is
the cost of the best strategy for Alice.
}

We present an alternate geometric view of $\pid$, which
generalizes naturally to other settings.
A subcube $C$ of $\zo^d$ is the set of points obtained by fixing some subset $S
\subseteq [d]$ coordinates to values in ${0,1}$. The
sparsity of $\T$ in a subcube $C$ is $\rho_{0,1}(\T, C) = |C|/|C \cap
\T|$. The notation $C \ni x$ means that $C$ contains $x$, hence
$\min_{C \ni x}$ is the minimum over all $C$ that contain $x$. One can show that for $x \in \T$,
$\max_{C \ni x}\rho_{0,1}(\T, C) = 2^{d - \pid(x, \T)}$, see appendix \ref{app:proof} for a proof.
This characterization motivates using $2^{-\pid(x,  \T)}$ as an
anomaly score: anomalies are points that lie in relatively sparse
subcubes. Low scores come with a natural witness:
a sparse subcube $\PID(x, \T)$ containing relatively few points from $\T$.

\eat{
{\bf Algorithmic justification: } Computing $\id(x, \T)$ is as hard as Set Cover, and is
thus NP-complete. On the other hand, since $\pid(x, \T) \leq \log(n)$,
this implies that optimal $\PID$s can be found in quasi-polynomial
time by only considering subsets $S$ of size $O(\log(n))$, and is
unlikely to be NP-complete (we do not know whether it is in P). Further, the use
of $\log_2(|\Imp|)$ as the penalty function lends itself to a natural
and easy-to-implement greedy algorithm: Start with $S$ being the empty
set. Pick coordinate $i \not\in S$. If $x_i \in \zo$ is the minority
value for the $i^{th}$ coordinate in $\Imp(x, \T, S)$, add $i$ to
$S$. Since this cuts $\Imp(x, \T, S)$ by half, it decreases the cost
function. If no such $i$ exists, then we terminate.
}

%%%%%%%%%%%%%%%%%%%%%%%%%%%%%%%%%%%%%%%%%%%%%%%%%%%%%%%%%%%%%%%%%%%%%%%%%%%%%%%%%%%%%%%%%
%%%%     Real-valued Setting
%%%%%%%%%%%%%%%%%%%%%%%%%%%%%%%%%%%%%%%%%%%%%%%%%%%%%%%%%%%%%%%%%%%%%%%%%%%%%%%%%%%%%%%%%

\subsection{The continuous setting}
\label{sec:cont}

Now assume that all the coordinates are real-valued, and
bounded. Without loss of generality, we may assume that they lie in
the range $\intzo$, hence $\T$ is a collection of $n$ points from $\intzo^d$.
An interval $I = [a,b], 0 \leq a \leq b \leq 1$ is of length $\len(I) = b -a$.
A subcube $C$ is specified by a subset of co-ordinates $S$ and intervals $I_j$ for
each $j \in S$. It consists of all points such that $x_j \in I_j$ for
all $j \in S$. To simplify our notation, we let $C$
be $I_1 \times I_2 \cdots \times I_d$ where $I_j = \intzo$ for $j \not\in
S$. Note that $\vol(C) = \Pi_{j}\len(I_j)$.
Define the sparsity of $\T$ in $C$ as $\rho(\T, C) = \vol(C)/|C \cap \T|$.
$\pidsc(x, T)$ is the maximum sparsity over all subcubes of $\intzo^d$  containing
$x$.

\begin{Def}
\label{def:pid}
For $x \in \T$, let
  \[ \PID(x, \T) = \arg\max_{C \ni x} \rho(\T, C), \ \ \pidsc(x, \T) =  \max_{C \ni x} \rho(\T,  C). \]
\end{Def}

To see the analogy to the Boolean case, define $\pid(x, \T) = -\log(\pidsc(x, \T))$.
Fix $C = \PID(x, \T)$. Since $\vol(C) = \prod_{j \in [d]}\len(I_j)$, we
can write
\begin{align}
  \label{eq:int-def}
  \pid(x, \T) =  \log(|C \cap \T|/\vol(C)) = \sum_{j \in [d]}\log(1/\len(I_j)) + \log(|C \cap \T|).
\end{align}
This exposes the similarities to Equation \eqref{eq:zo-def2}: $C \cap \T$ is exactly the set of impostors for $x$, whereas $\sum_{j
  \in [d]}\log(1/\len(I_j))$ is the analog of $|S|$. In the boolean setting, we pay $1$ for each coordinate from
  $S$, here the cost ranges in $[0, \infty)$ depending on the length of the interval.
In the continuous setting, the $j \not\in S$ iff $I_j = \intzo$
  hence $\log(1/\len(I_j)) = 0$, hence we pay nothing for coordinates outside $S$.
Restricting to an interval of length $p$ costs
  $\log(1/p)$. If $p = 1/2$, we pay $1$, which is analogous to the
  Boolean case where we pay $1$ to cut the domain in half. This
  addresses the issue of having to pay more for higher precision. Note
  also that the definition is \emph{scale-invariant} as multiplying a
  coordinate by a constant changes the volume of all subcubes by the
  same factor.

  {\bf Other attributes: } To handle attributes over a domain $D$, we
  need to specify what subsets of $D$ are intervals and how we measure
  their length.
  For discrete attributes, it is natural to define $\len(I) = |I|/|D|$. When the
  domain is  ordered intervals are naturally defined, for instance {\em months between
  April and September} is an interval of length $1/2$. We could also
  allow wraparound in intervals, say {\em months between November and
  March}. For unordered discrete values, the right definition of
  interval could be singleton sets, like {\em country =
    Brazil} or certain subsets, like {\em continent = the
    Americas}. The right choice will depend on the dataset.
  Our definition is flexible enough to handle this:
  We can make independent choices for each coordinate, subcubes and
  volumes are then defined as products, and $\pidsc$ can be defined
  using definition \ref{def:pid}.

{\bf IDs and PIDs.}
The notion of IDs for a point is natural and has been studied in the
computational learning literature under various names: the teaching
dimension of a hypothesis class \cite{GoldmanK95}, discriminant \cite{Natarajan91},
specifying set \cite{AnthonyBCS92} and witness set
\cite{KushilevitzLRS96}. Our work is inspired by the work of
Wigderson and Yehudayoff \cite{WigdersonY16} on population recovery,
which is the task of learning mixtures of certain discrete
distributions on the Boolean cube. Their work introduces the notion of
partial identification in the Boolean setting.  The terminology of IDs
and impostors is from their work. They also consider PIDs,
but with a different goal in mind (to minimize the depth of a certain
graph constructed using the PID relation).
Our definition of $\pid$, the extension of partial identification to
the continuous setting  and its application to anomaly detection are
new contributions.

\newcommand{\numSamp}{\mathrm{numSamples}}
\newcommand{\degree}{\mathrm{degree}}
\newcommand{\depth}{\mathrm{depth}}

\section{The PIDForest algorithm}
\label{sec:pidf}

%\Revtwo{This section is a bit hard to follow.}

We do not know how to compute $\pidsc$ exactly, or even a provable approximation of it in a way that scales well with both $d$ and $n$.  The \pidF~algorithm described below is
heuristic designed to approximate $\pidsc$.
Like with \iF, the \pidF ~algorithm builds an ensemble of decision trees, each tree
is built using a sample of the data set and partitions the space into
subcubes. However, the way the trees are constructed and the criteria
by which a point is declared anomalous are very different.
Each node of a tree corresponds to a subcube $C$, the children of $C$ represent
a disjoint partition of $C$ along some axis $i \in [d]$ (\iF\ always
splits $C$ into two , here we allow for a finer partition).
The goal is to have large variance in the sparsity of the subcubes. Recall that the sparsity of a subcube $C$ with respect to a data set $\T$ is $\rho(C, \T) =
\vol(C)/|C \cap \T|$. Ultimately, the leaves with large $\rho$ values
will point to regions with anomalies.

For each tree, we pick a random sample $\P \subseteq \T$ of $m$ points, and
use that subset to build the tree.
Each node $v$ in the tree corresponds to subcube $C(v)$, and a set of points $P(v)
= C(v) \cap \P$. For the root, $C(v) = \intzo^d$ and $P(v) = \P$.
At each internal node, we pick a coordinate $j \in [d]$,
and breakpoints $t_1 \leq \cdots \leq t_{k-1}$ which partition $I_j$ into
$k$ intervals, and split $C$ into $k$ subcubes. The number of
partitions $k$ is a hyper-parameter (taking $k < 5$ works well in practice). We then partition the
points $P(v)$ into these subcubes. The partitions stop when the tree reached some specified maximum depth or when  $|P(v)| \leq 1$.  The key algorithmic problem is how to choose
the coordinate $j$ and the breakpoints by which it should be
partitioned. Intuitively we want to partition the cube
into some sparse regions and some dense regions. This intuition is formalized next.

Let $I_j \subseteq \intzo$ be the projection of $C$ onto coordinate
$i$. Say the breakpoints are chosen so that we partition $I_j$
into $I_j^1$, \ldots, $I_j^k$. This partitions $C$ into $C^1, \ldots, C^k$
where the intervals corresponding to the other coordinates stay the same.
We first observe that in any partition of $C$, the {\em average} sparsity of the
subcubes when weighted by the number of points is the same.
Let
\[ p_i \defeq \frac{\len(I_j^i)}{\len(I)} = \frac{\vol(C^i)}{\vol(C)},  q_i \defeq \frac{|P
\cap C^i|}{|P|}.\]
Hence
\[ \rho(C^i) = \frac{\vol(C^i)}{|P \cap C^i|} =  \frac{p_i
  \vol(C)}{q_i |P|} = \frac{p_i\rho(C)}{q_i}.\]

\eat{
Let
\[ p_i \defeq \len(I_j^i)/\len(I) = \vol(C^i)/\vol(C), q_i \defeq |P
\cap C^i|/|P|.\]
Hence
\[ \rho(C^i) = \vol(C^i)/|P \cap C^i| =  (p_i \vol(I))/(q_i |P|) = (p_i\rho(C))/q_i.\]
}

Since a $q_i$ fraction of points in $P$ have sparsity $\rho(C_i)$, the
expected sparsity for a randomly chosen point from $P$ is
\[ \sum_{i \in [k]} q_i \rho(C^i) = \sum_{i \in [k]} p_i \rho(C) = \rho(C). \]
In words, in any partition of $C$ if we pick a point randomly from
$P(v)$ and measure the sparsity of its subcube, on expectation  we get
$\rho(C)$. Recall that our goal is to split $C$ into sparse and dense subcubes.
Hence a natural objective is to maximize the \emph{variance} in the
sparsity:
\begin{align}
\label{eq:var-pk}
  \Var(\P,\{C^i\}_{i \in [k]}) = \sum_{i \in [k]} q_i (\rho(C^i) - \rho(C))^2 =
  \sum_{i \in [k]}q_i\rho(C^i)^2 - \rho(C)^2.
\end{align}

A partition that produces large variance in the sparsity needs to
partition space into some sparse regions and other dense regions,
which will correspond to outliers and normal regions
respectively. Alternately, one might choose partitions to optimize the
maximum sparsity of any interval in the partition, or some higher moment
of the sparsity. Maximizing the variance has the advantage that it
turns out to equivalent to a well-studied problem about histograms,
and admits a very efficient streaming algorithm. We continue splitting
until a certain predefined  depth is reached, or points are
isolated. Each leaf is labeled with the  sparsity of its subcube.

%%%%%%%%%%%%%%%%%%%%%%%%%%%%%%%%%%%%
%%% Space saving dirty trick     %%%
%%%%%%%%%%%%%%%%%%%%%%%%%%%%%%%%%%%%

\eat{
 Let
\[ \frac{\len(I_j^i)}{\len(I)} = \frac{\vol(C^i)}{\vol(C)} =  p_i.\]
Let $|P \cap C^i|/|P| = q_i$, so that
\[ \rho(C^i) = \frac{\vol(C^i)}{|P \cap C^i|} =  \frac{p_i
  \vol(I)}{q_i |P|} = \rho(C)\frac{p_i}{q_i}.\]
Since a $q_i$ fraction of points in $P$ have sparsity $\rho(C_i)$, the
expected sparsity for a randomly chosen point from $P$ is
\[ \sum_j q_j \rho(C_j) = \sum_j p_j \rho(C) = \rho(C). \]
So indeed the average sparsity across points stays the same. Recall, our goal is to split $C$ into subcubes that
are sparse and subcubes that are dense. Hence a natural objective is to
maximize the \emph{variance} in the sparsity of a random point in $P$:
\begin{align}
\label{eq:var-pk}
  \Var(\P, \{C^i\}_{i \in [k]}) = \sum_j q_j (\rho(C_j) - \rho(C))^2 =
  \sum_jq_j\rho(C_j)^2 - \rho(C)^2
\end{align}
}

\myalgo{\pidF\ Fit}{

  Params: Num of trees $t$, Samples $m$, Max degree $k$, Max depth $h$.\\
  Repeat $t$ times:\\
  $~~~$ Create root node $v$.\\
  $~~~$ Let $C(v) = \intzo^d$, $P(v) \subseteq \T$ be
  a random subset of size $m$ .\\
  $~~~$ $\mathrm{Split}(v)$\\
  \\
  $\mathrm{Split}(v)$: \\
  $~~~$ For $j \in [d]$, compute the best split into $k$ intervals.\\
  $~~~$ Pick $j$ that maximizes variance, split $C$ along $j$ into $\{C^i\}_{i=1}^k$.\\
  $~~~$ For $i \in [k]$ create child $v_i$ s.t. $C(v_i) = C^i, P(v_i) = P(v) \cap C^i$.\\
  $~~~$ If $\depth(v_i) \leq h$ and $|P(v_i)| > 1$ then
  $\mathrm{Split}(v_i)$. \\
  $~~~$ Else, set $\pidsc(v_i) = \vol(C(v_i))/|P(v_i)|$.
  }

In Appendix \ref{sec:splitting}, we show that the problem of finding
the partition along a coordinate $j$ that maximizes the variance in
the sparsity  can be reduced to the problem of finding a $k$-histogram
for a discrete function $f:[n] \rgta \R$, which minimizes the squared
$\ell_2$ error. This is  a well-studied problem
\cite{koudas2000optimal,gilbert2002fast,guha2002histogramming} and
there  an efficient one-pass streaming algorithm for computing
near-optimal histograms due to Guha {\em et al.} \cite{GuhaKS06}. We
use this algorithm to compute the best split along each coordinate,
and then choose the coordinate that offers the most variance reduction.
Using their algorithm, finding the optimal split for a node
takes time $O(dm\log(m))$. This is repeated at most $k^h$ times for
each tree (typically much fewer since the trees we build tend to be
unbalanced), and $t$ times to create the forest. We typically choose
$m \leq 200$, $k \leq 5$, $h \leq 10$ and $t \leq 50$.

Producing an anomaly score for each point is
fairly straightforward. Say we want to compute a score for $y \in
\intzo^d$. Each tree in the forest maps $y$ to a leaf node $v$
and gives it a score $\pidsc(v)$. We take the 75\% percentile
score as our final score (any robust analog of the max
will do).

\subsection{Comparison to Isolation Forest}\label{sec:if}
 \iF\ repeatedly samples a set
$S$ of $m$ points from $\T$ and builds a random tree with those points
as leaves. The tree is built by choosing a random co-ordinate $x_i$, and a random value in its range
about which to split. The intuition is that anomalous points will be
easy to separate from the rest, and will be isolated at small depth.
What kind of points are likely to be labeled anomalous by \iF?

\eat{
Since we care about small depth points, we do not need to isolate {\em normal} points, so
we stop the tree at depth $\log(m)$. For a point $x$, its average depth in the
ensemble of trees is its anomaly score, points with small
average depth are likely to be anomalies.
}

In one direction, if a point is isolated at relatively low depth $k$ in a tree,
then it probably belongs in a sparse subcube. Indeed, a node at depth
$k$ corresponds to a subcube $C$ of expected volume $2^{-k}$, which is
large for small $k$. The fact that the sample contains no points from
$C$ implies that $C \cap \T$ is small, with high probability (this can be
made precise using a VC-dimension argument). Hence $\rho(C, T) =
\vol(C)/|C \cap \T|$ is fairly large.

Being in a sparse subcube is necessary but not sufficient.
This is because \iF\ chooses which coordinate we
split on as well as the breakpoint at random.  Thus to be isolated at
small depth frequently, a point needs to lie in an \emph{abundant} number of low-density
subspaces: picking splits at random should have a good chance of defining such a subspace.
Requiring such an abundance of sparse subcubes can be
problematic. Going back to the animals example, isolating white
elephants is hard unless both Color and Type are used as attributes,
as there is no shortage of elephants or white animals.
Moreover, which attributes are relevant can
depend on the point: weight might be irrelevant in
isolating a white elephant, but it might be crucial to isolating a
particularly large elephant. This causes \iF\ to perform poorly in the presence of
irrelevant attributes, see for instance \cite{BandaragodaTALZ18}.

This is the fundamental difference between \pidF\
and \iF\ and its variants. \density zooms in on coordinates with
signal---where a split is most beneficial.
Attributes with little signal are unlikely to be chosen for splitting. For concrete
examples, see Section \ref{sec:synthetic}. The tradeoff is that we incur a
slightly higher cost at training time, the cost of prediction stays
pretty much the same.

\section{Real-world Datasets}
\label{sec:real}

We show that \pidF\ performs favorably in comparison to several
popular anomaly detection algorithms on  real-world benchmarks. We
select datasets from varying domains, and with different number of
datapoints, percentage of anomalies and dimensionality. The code and
data for all experiments is available online.\footnote{\url{https://github.com/vatsalsharan/pidforest}}

%\Revtwo{Are these standard datasets?}
{\bf Dataset Descriptions: }
The first set of datasets are classification datasets from the UCI
\citep{asuncion2007uci} and openML repository
\citep{OpenML2013}. Three of the datsets---\emph{Thyroid},
\emph{Mammography} and \emph{Seismic}---are naturally suited to
anomaly detection as they are binary classification tasks where one of
the classes has a much smaller occurrence rate (around $5 \%$) and
hence can be treated as anomalous. \emph{Thyroid} and
\emph{Mammography} have real-valued attributes whereas \emph{Seismic}
has categorical attributes as well. Three other
datasets---\emph{Satimage-2}, \emph{Musk} and \emph{Vowels}---are
classification datasets with multiple classes, and we combine the
classes and divide them into inliers and outliers as in
\cite{aggarwal2015theoretical}. Two of the datasets---\emph{http} and
\emph{smtp}---are derived from the KDD Cup 1999 network intrusion
detection task and we preprocess them as in
\cite{yamanishi2004line}. These two datasets have have significantly
more datapoints (about 500k and 100k respectively) and a smaller
percentage of outliers (less than $0.5\%$).

%\Revtwo{How would the anomaly detection algorithms compare with the
%performance of a supervised method on these tasks?}

The next set of real-world datasets---\emph{NYC taxicab}, \emph{CPU
  utilization}, \emph{Machine temperature (M.T.)} and \emph{Ambient
  temperature (A.T.)}---are time series datasets from the Numenta anomaly
detection benchmark  which have been hand-labeled with anomalies rooted in
real-world causes \cite{ahmad2017unsupervised}. The length of the time series is 10k-20k,
with about $10\%$ of the points marked anomalous. We use the
standard technique of \emph{shingling} with a sliding window of width
$10$, hence each data point becomes a $10$ dimensional vector of
$10$ consecutive measurements from the time series.  Detailed
parameters of the datasets are given in Table \ref{tab:datasets}.

\begin{table*}
	\centering
	\begin{tabular}{@{}llll@{}}
		\toprule
		\multicolumn{1}{l}{Data set} & \multicolumn{1}{l}{$n$} & \multicolumn{1}{l}{$d$} & \multicolumn{1}{l}{$\#$outliers ($\%$)}  \\ \toprule

		Thyroid & 7200  & 6 & 534 (7.42$\%$) \\ \midrule
		Mammography (Mammo.)& 11183  & 6 & 250 (2.32$\%$) \\ \midrule
		Seismic & 2584  & 15 & 170 (6.5$\%$) \\ \midrule
		Satimage-2 & 5803  & 36 & 71 (1.2$\%$) \\ \midrule
		Vowels & 1456  & 12 & 50 (3.4$\%$) \\ \midrule
		Musk & 3062  & 166 & 97 (3.2$\%$)\\ \midrule
		http & 567479  & 3 & 2211 (0.4$\%$)\\ \midrule
		smtp & 95156  & 3 & 30 (0.03$\%$)\\ \midrule
		NYC taxicab & 10321  & 10 & 1035 (10$\%$) \\ \midrule
		Ambient Temperature (A.T.) & 7267  & 10 & 726 (10$\%$) \\ \midrule
		CPU utilization & 18050  & 10 & 1499 (8.3$\%$) \\ \midrule
		Machine temperature (M.T.) & 22695  & 10 & 2268 (10$\%$) \\ \bottomrule

	\end{tabular}
	\caption{Details of real-world datasets. The first 8 datasets are derived from classification tasks, and the last 4 are from time series with known anomalies.}
	\label{tab:datasets}
\end{table*}

{\bf Methodology: } We compare \pidF with six popular anomaly
detection algorithms: Isolation Forest (iForest), Robust Random Cut
Forest (RRCF), one-class SVM (SVM), Local Outlier Factor (LOF),
k-Nearest Neighbour (kNN) and Principal Component Analysis (PCA). We
implement \pidF\ in Python, it takes about 500 lines of code.
For iForest, SVM and LOF we used the scikit-learn implementations, for kNN
and PCA we used the implementations on PyOD \citep{zhao2019pyod} , and
for RRCF we use the implementation from \citep{Tharindu}.  Except for
RRCF, we run each algorithm with the default hyperparameter setting as
varying the hyperparameters from their default values
did not change the results significantly. For RRCF, we use $500$ trees
instead of the default $100$ since it yielded significantly better
performance. For PIDForest, we fix the hyperparameters of depth to $10$,
number of trees to $50$, and the number of samples used to build each
tree to $100$. We use the area under the ROC curve (AUC) as the
performance metric. As \iF, \pidF\ and RRCF are randomized, we repeat these algorithms for 5
runs and report the mean and standard deviation. SVM, LOF, kNN and PCA
are deterministic, hence we report a single AUC number for them.

{\bf Results:} We report the results in Table
\ref{tab:results}. \pidF\ is the top performing or jointly top
performing algorithm in 6 out of the 12 datasets, and iForest, kNN and
PCA are top performing or jointly top performing algorithms in 3
datasets each. Detailed ROC performance curves of the algorithms are
given in Fig. \ref{fig:roc1} and \ref{fig:roc2} in the Appendix. While the running
time of our fit procedure is slower than \iF, it is comparable to RRCF
and faster than many other methods. Even our vanilla Python
implementation on a laptop computer only takes about 5 minutes to fit
a model to our largest dataset which has half a million points.

Recall from Section \ref{sec:pidf} that
\pidF\ differs from \iF\ in two ways, it optimizes for
the axis to split on, and secondly, it uses sparsity instead of depth
as the anomaly measure.
To further examine the factors which contribute to the favorable
performance of \pidF, we do an ablation study through two
additional experiments.

%We can see that both iForest and PIDForest perform significantly better than LOF and OCSVM on 9 out of 12 datasets. Among iForest and PIDForest, PIDForest is the top performing algorithm in 5 out of the 12 datasets, iForest is the top algorithm in 3 out of the 12 datasets, with both of them being the best algorithm on one dataset.

\begin{table*}
	\centering
	\begin{tabular}{@{}llllllll@{}}
		\toprule
		\multicolumn{1}{c}{Data set} & \multicolumn{1}{c}{PIDForest} & \multicolumn{1}{c}{iForest}  & \multicolumn{1}{c}{RRCF} & LOF & SVM &  kNN & PCA \\ \toprule

		Thyroid & \textbf{0.876 $\pm$ 0.013} &0.819 $\pm$ 0.013 & 0.739$\pm$ 0.004 & 0.737 & 0.547 &0.751 & 0.673 \\ \midrule
		Mammo. & 0.840 $\pm$ 0.010 &0.862 $\pm$ 0.008 & 0.830 $\pm$ 0.002 & 0.720 & 0.872 &0.839 & \textbf{0.886}\\ \midrule
		Seismic & \textbf{0.733 $\pm$ 0.006} &0.698 $\pm$ 0.004 & 0.701 $\pm$ 0.004 & 0.553 & 0.601 & \textbf{0.740} & 0.682\\ \midrule
		Satimage & 0.987 $\pm$ 0.001 & \textbf{0.994 $\pm$ 0.001}& \textbf{0.991 $\pm$ 0.002} & 0.540 & 0.421 &0.936 & 0.977\\ \midrule
		Vowels & 0.741 $\pm$ 0.008 &0.736 $\pm$ 0.026 & 0.813$\pm$ 0.007 & 0.943 & 0.778 & \textbf{0.975} & 0.606\\ \midrule
		Musk & \textbf{1.000 $\pm$ 0.000} & \textbf{0.998 $\pm$ 0.003}& 0.998 $\pm$ 0.000 & 0.416 & 0.573 &0.373 & \textbf{1.000}\\ \midrule
		http & 0.986 $\pm$ 0.004 & \textbf{1.000 $\pm$ 0.000} & 0.993 $\pm$ 0.000 & 0.353 & 0.994 &0.231 & 0.996\\ \midrule
		smtp & \textbf{0.923 $\pm$ 0.003} &0.908 $\pm$ 0.003 & 0.886 $\pm$ 0.017 & 0.905 & 0.841 &0.895 & 0.823 \\ \midrule
		NYC  & 0.564 $\pm$ 0.004 &0.550 $\pm$ 0.005 & 0.543 $\pm$ 0.004 & 0.671 & 0.500 & \textbf{0.697} & 0.511\\ \midrule
		A.T. & \textbf{0.810} $\pm$ 0.005 &0.780 $\pm$ 0.006 & 0.695 $\pm $0.004 & 0.563 & 0.670 &0.634 & 0.792\\ \midrule
		CPU & \textbf{0.935 $\pm$ 0.003} &0.917 $\pm$ 0.002 & 0.785 $\pm$ 0.002 & 0.560 & 0.794 &0.724 & 0.858\\ \midrule
		M.T. & 0.813 $\pm$ 0.006 &0.828 $\pm$ 0.002 & 0.7524 $\pm$ 0.003 & 0.501 & 0.796 &0.759 & \textbf{0.834}\\ \bottomrule

	\end{tabular}
	\caption{Results on real-world datasets.
		We bold the algorithm(s) which get the best AUC.}
	\label{tab:results}
\end{table*}

{\bf Choice of split:} Optimizing for the choice of split rather than choosing one at random
seems valuable in the presence of irrelevant
dimensions. To measure this effect, we added $50$ additional random
dimensions sampled uniformly in the range $[0,1]$ to two
low-dimensional datasets from Table \ref{tab:results}---\emph{Mammography}
and \emph{Thyroid} (both datasets are $6$ dimensional).
In the \emph{Mammography} dataset, PIDForest (and many other
algorithms as well) suffers only a small $2\%$ drop in performance,
whereas the performance of iForest drops by $15\%$. In
the \emph{Thyroid} dataset, the performance of all algorithms drops
appreciably. However, \pidF\ has a $13\%$ drop in performance, compared to a $20\%$ drop for
\iF.  The detailed results are given in Table \ref{tab:results_noisy}.

\begin{table*}
	\centering
	\begin{tabular}{@{}llllllll@{}}
		\toprule
		\multicolumn{1}{c}{Data set} & \multicolumn{1}{c}{PIDForest} & \multicolumn{1}{c}{iForest}  & RRCF & LOF & SVM &  kNN & PCA \\ \toprule
		Thyroid$^{*}$ & \textbf{0.751 $\pm$ 0.035} &0.641 $\pm$ 0.023 & 0.530 $\pm$ 0.005 & 0.492 & 0.494 &0.495 & 0.614\\ \midrule
		Mammography$^{*}$ & 0.829 $\pm$ 0.016 &0.722 $\pm$ 0.016 & 0.797 $\pm$ 0.013 & 0.628 & \textbf{0.872} &0.817 & 0.768\\ \bottomrule

	\end{tabular}
	\caption{For the first two datasets in Table \ref{tab:results} we add 50 noisy dimensions to examine the performance of algorithms in the presence of irrelevant attributes. We bold the algorithm(s) which get the best AUC, up to statistical error.}
	\label{tab:results_noisy}
\end{table*}

{\bf Using sparsity instead of depth:} In this experiment, we test the
hypothesis that the sparsity of the leaf is a better anomaly score
than depth for the \pidF\ algorithm. The performance of \pidF\ deteriorates noticeably
with depth as the score, the AUC for \emph{Thyroid} drops to $0.847$
from $0.876$, while the AUC for \emph{Mammography} drops to $0.783$
from $0.840$.

\section{Synthetic Data}\label{sec:synthetic}
%\Revtwo{There could be two anomaly detection setups, i) the batch setup where you have a bunch of anomalous and non-anomalous points and want to find anomalies, versus ii) a setup where we have only non-anomalous data to train, then get a single new point, and have to decide if it is an anomaly. Does our algorithm work in the second? We don't talk about the second setup at all right now.}

We compare \pidF\ with popular anomaly detection algorithms on
synthetic benchmarks. The first experiment checks how
the algorithms handle duplicates in the data. The second experiment
uses data  from a mixture of Gaussians, and highlights the importance of
the choice of coordinates to split in \pidF. The third experiment tests the ability of the algorithm to detect anomalies in
time-series. In all these experiments, \density outperforms prior art. We also evaluate the robustness of PIDForest to the choice of hyperparameters in the fourth experiment, and find its performance to be robust to a wide range of hyperparameter choices. 

% We compare the performance of \density to that of popular anomaly detection algorithms over various synthetic benchmarks. The goal is to demonstrate how it differs from prior work in a more systematic way. The first set of experiments checks how well the algorithm handles duplicates in the data, specifically whether duplicates make it harder to tune the parameters. The first set of experiments highlights the importance of splitting coordinates that reduce the variance. We show this by adding dimensions that have a uniform distribution (i.e. do not change the likelihood of points). The third set of experiments tests the ability of the algorithm to detect anomalies in a time-series which is a very common use-case and one for which it is known that \iF\ does well. In all these experiments \density outperforms prior art.
%In the experiments the forest was built from samples of the data, the full data set was then fed to the model to produce the anomalies.

\subsection{Masking and Sample Size}
\label{sec:masking}
\begin{figure}

		\centering
		\includegraphics[width=0.5\textwidth]{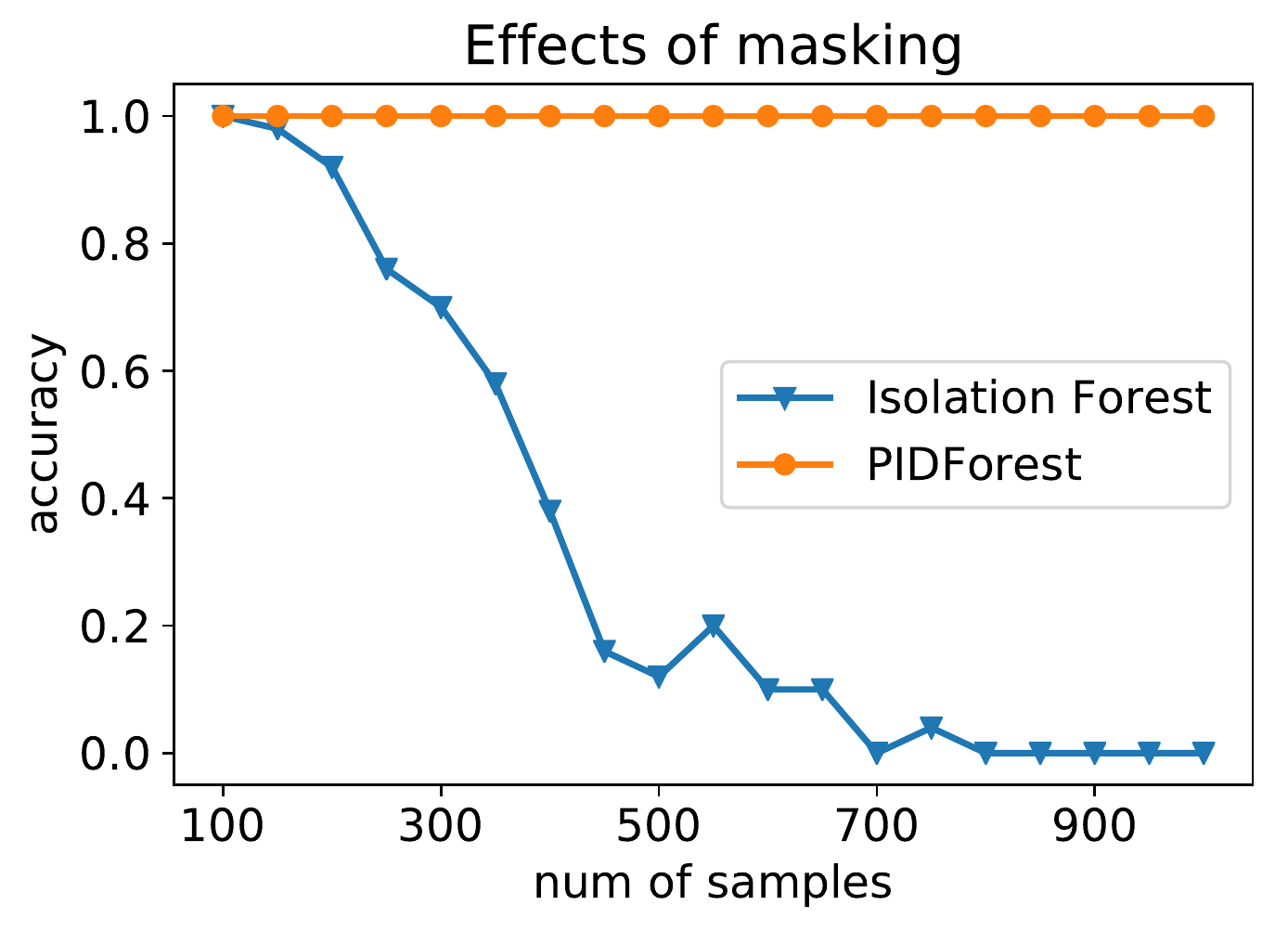}
		\caption{Masking. Accuracy is
			the fraction of true outliers in the top
			$5\%$ of reported anomalies. }
		\label{fig:masking}

%	\caption{Synthetic experiments on masked anomalies.}
\end{figure}

It is often the case that anomalies repeat multiple
times in the data. This phenomena is called \emph{masking} and is a
challenge for many algorithms. \iF\ counts on sampling to counter
masking: not too many repetitions occur in the sample. But the
performance is sensitive to the sampling rate, see
\cite{GuhaMRS16,BandaragodaTALZ18}. To demonstrate it, we create a
data set of $1000$ points in $10$ dimensions. $970$ of these points
are sampled randomly in $\{-1,1\}^{10}$ (hence most of these points
are unique). The remaining $30$ are the all-zeros vector, these constitute a masked anomaly.
We test if the zero points are declared as anomalies by
\pidF\ and \iF\ under varying sample sizes. The
results are reported in Fig.~\ref{fig:masking}.
Whereas \pidF\ consistently reported these points as anomalies, the performance of
\iF\ heavily depends on the sample size. When it is small,
then masking is negated and the anomalies are
caught, however the points become hard to isolate when the
sample size increases.

\eat{In particular, for Isolation Forest,
when a forest is being constructed from samples of the data, the size
of the sample determines how  frequently each point will be  in the
sample and this may affect the result.
 how much each point will be isolated in the sample}

%While Local Outlier Factor is not reported in the graph, it should be reported that failed to report any of the zero points until the precision is unrealistically low.

\subsection{Mixtures of Gaussians and Random Noise}\label{sec:gaussian}
\begin{figure}
	\begin{subfigure}{0.45\textwidth}
		\centering
		\includegraphics[width=\textwidth]{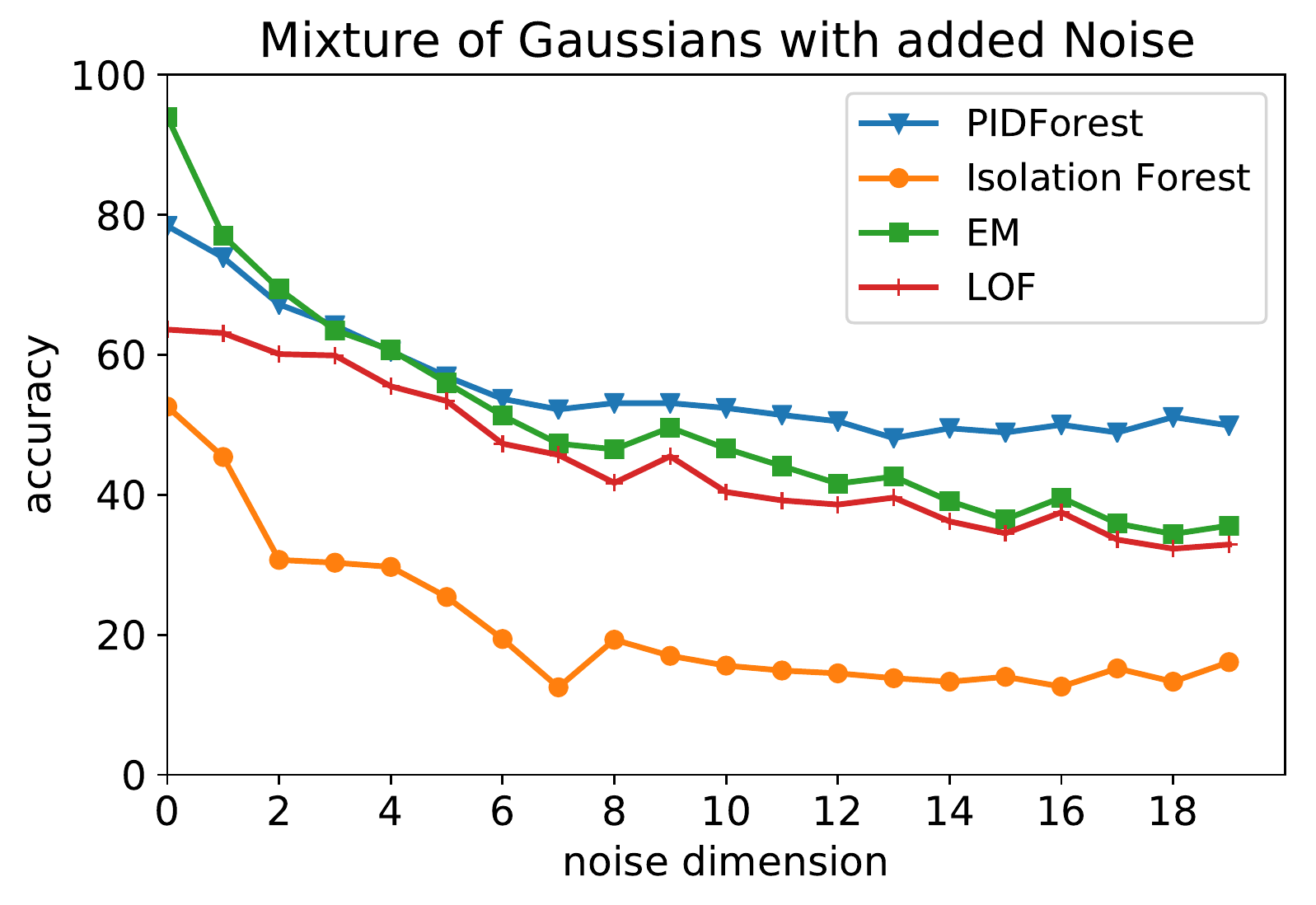}
		\caption{Comparison of PIDForest with Isolation Forest, EM and LOF.}
		\label{fig:noisy_Gaussian1}
	\end{subfigure}\hfill
	\begin{subfigure}{0.45\textwidth}
		\centering
		\includegraphics[width=\textwidth]{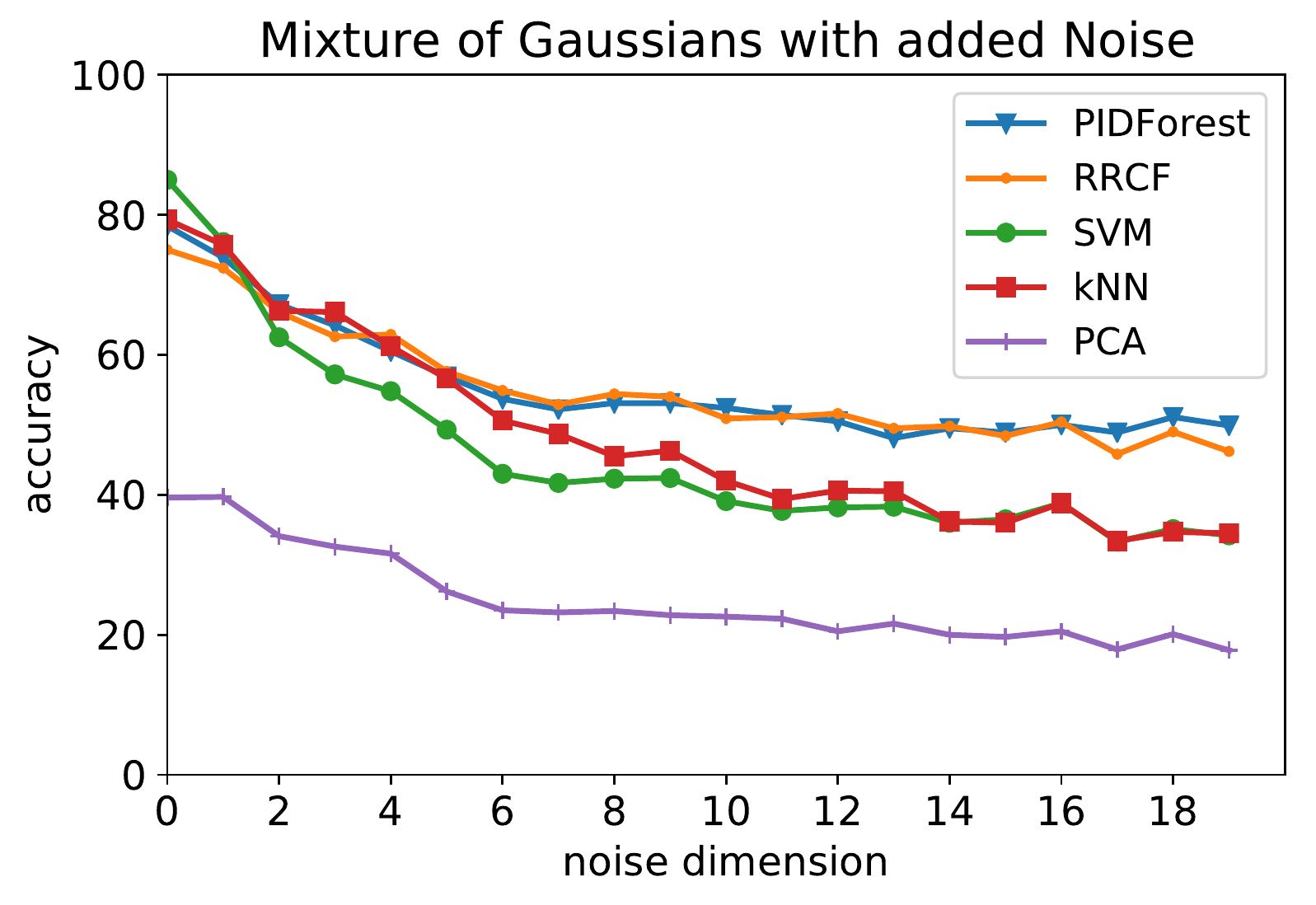}
		\caption{Comparison of PIDForest with RRCF, SVM, kNN and PCA. }
		\label{fig:noisy_Gaussian2}
	\end{subfigure}\hfill
	\caption{Synthetic experiments on Gaussian data (noisy dimensions are uniform in $[-2,2]$). For clarity, we split the results into two figures. $y-$axis measures how
		many of the $100$ true anomalies were reported by
		the algorithm in the top $100$ anomalies.}
\end{figure}

\begin{figure}
	\begin{subfigure}{0.45\textwidth}
		\centering
		\includegraphics[width=\textwidth]{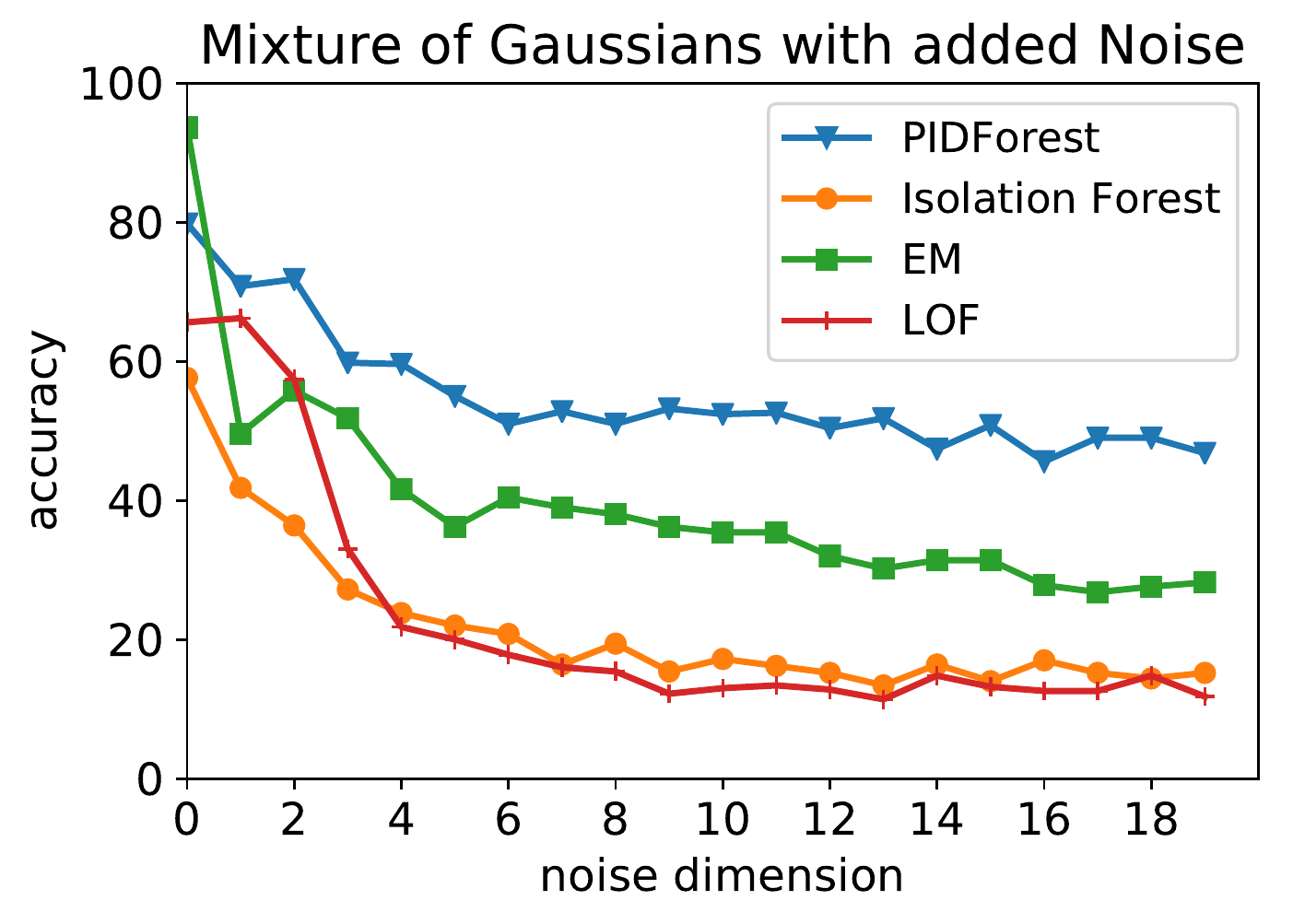}
		\caption{Comparison of PIDForest with Isolation Forest, EM and LOF. }
		\label{fig:noisy_Gaussian_morenoise1}
	\end{subfigure}\hfill
	\begin{subfigure}{0.45\textwidth}
		\centering
		\includegraphics[width=\textwidth]{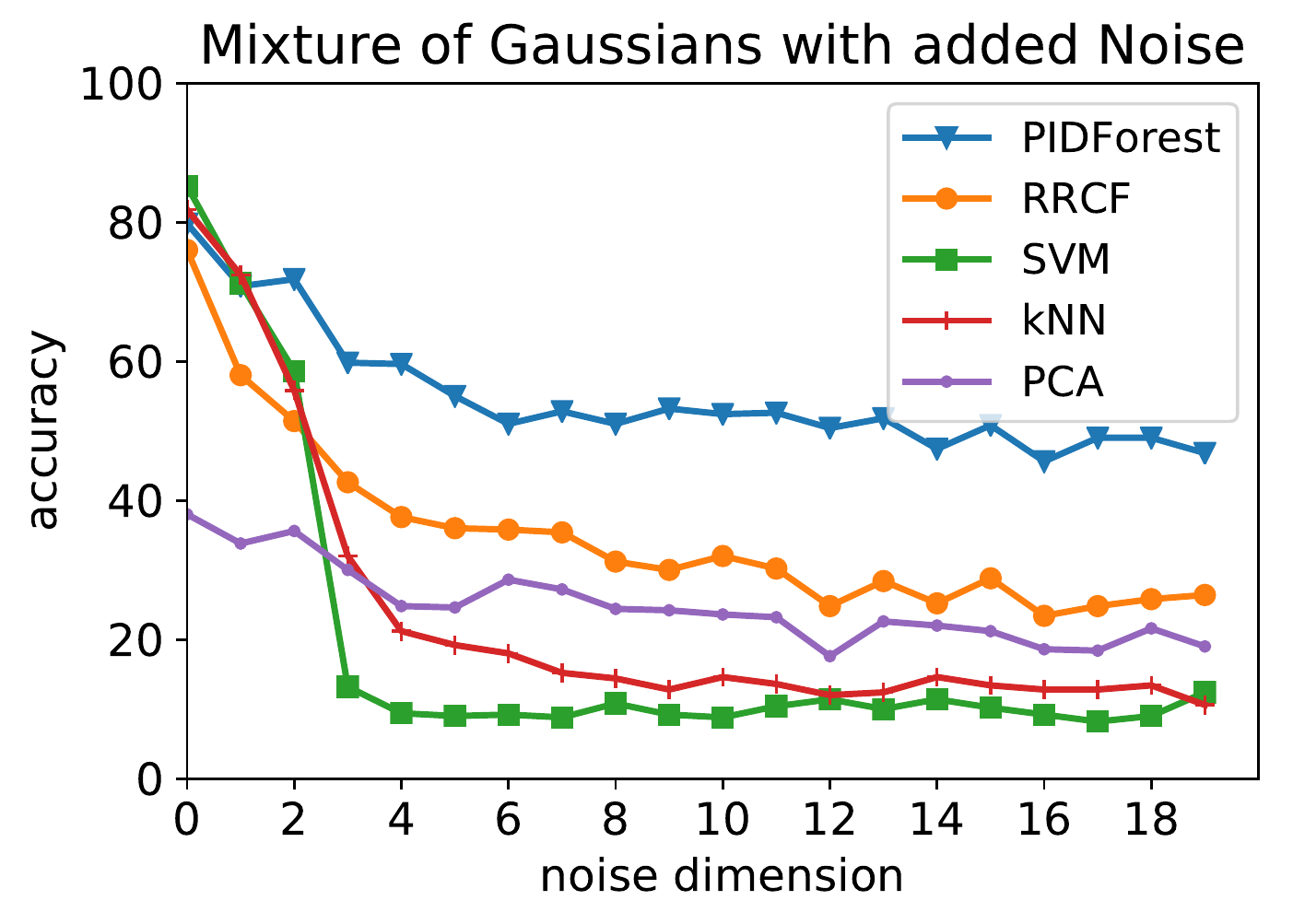}
		\caption{Comparison of PIDForest with RRCF, SVM, kNN and PCA. }
		\label{fig:noisy_Gaussian_morenoise2}
	\end{subfigure}\hfill
	\caption{Synthetic experiments on Gaussian data (noisy dimensions are uniform in $[-10,10]$). For clarity, we split the results into two figures. $y-$axis measures how
		many of the $100$ true anomalies were reported by
		the algorithm in the top $100$ anomalies.}
\end{figure}

We use a generative model where the ground truth anomalies are
the points of least likelihood. The first two coordinates of the $1000$
data points are sampled by taking a mixture of two $2-$dimensional
Gaussians with different means and covariances (the eigenvalues are
$\{1,2\}$, the eigenvectors are chosen at random).  The remaining $d$
dimensions are sampled uniformly in the range $[-2,2]$.  We run
experiments varying $d$ from $0$ to $19$. In each case we take the $100$ points
smallest likelihood points as the ground truth anomalies.
\eat{Note that
since the noisy dimensions are uniform the likelihood is completely
determined by the first two dimensions.}
For each algorithm we examine
the $100$ most anomalous points and calculate how many of these belong
to the ground truth. We compare \density with Isolation Forest, Local Outlier Factor (LOF), an algorithm that uses Expectation Maximization (EM) to fit the data to a mixture of
Gaussians, RRCF, SVM, kNN and PCA. The results are reported in Figs. \ref{fig:noisy_Gaussian1} and \ref{fig:noisy_Gaussian2}.
%Note that even without noise ($d=0$) \density among the best generic algorithm. As the number of noisydimensions increase \density focuses on the dimensions with signal, soit performs better. 
Some observations:
\begin{enumerate}
	\item  The performance of \iF\ degrades rapidly with $d$, once
          $d \geq 6$ it effectively outputs a random set. Noticeably,
          \pidF\ performs better than \iF\ even when $d=0$. This is
          mainly due to the points between the two centers being
          classified as normal by \iF. \pidF\ classifies them correctly
          as anomalous even though they are assigned to leaves that are deep in the tree.

	\item The EM algorithm is specifically designed to fit a mixture of two Gaussians, so it does best for
          small or zero noise, i.e. when $d$ is small. PIDForest
          beats it once $d>2$.

          %. As $d$ increases the data distribution is further away from a mixture. \density\ matches or improves once $d>2$.

	\item Apart from PIDForest, a few other algorithms such as RRCF, SVM and kNN also do  well---but their performance is crucially dependent on the fact that the noise is of the same scale as
	the Gaussians. If we change the scale of the noise (which
	could happen if the measuring unit changes), then the
	performance of all algorithms except PIDForest drops significantly. In Figs. \ref{fig:noisy_Gaussian_morenoise1} and \ref{fig:noisy_Gaussian_morenoise2}, we repeat the same experiment as above but with the noisy dimensions being uniform in $[-10,10]$ (instead of $[-2,2]$). The performance of PIDForest is essentially unchanged, and it does markedly better than all other algorithms.
\end{enumerate}

\subsection{Time Series} 
\begin{figure*}
	\centering
	\begin{subfigure}{0.45\textwidth}
		\centering
		\includegraphics[width=\textwidth]{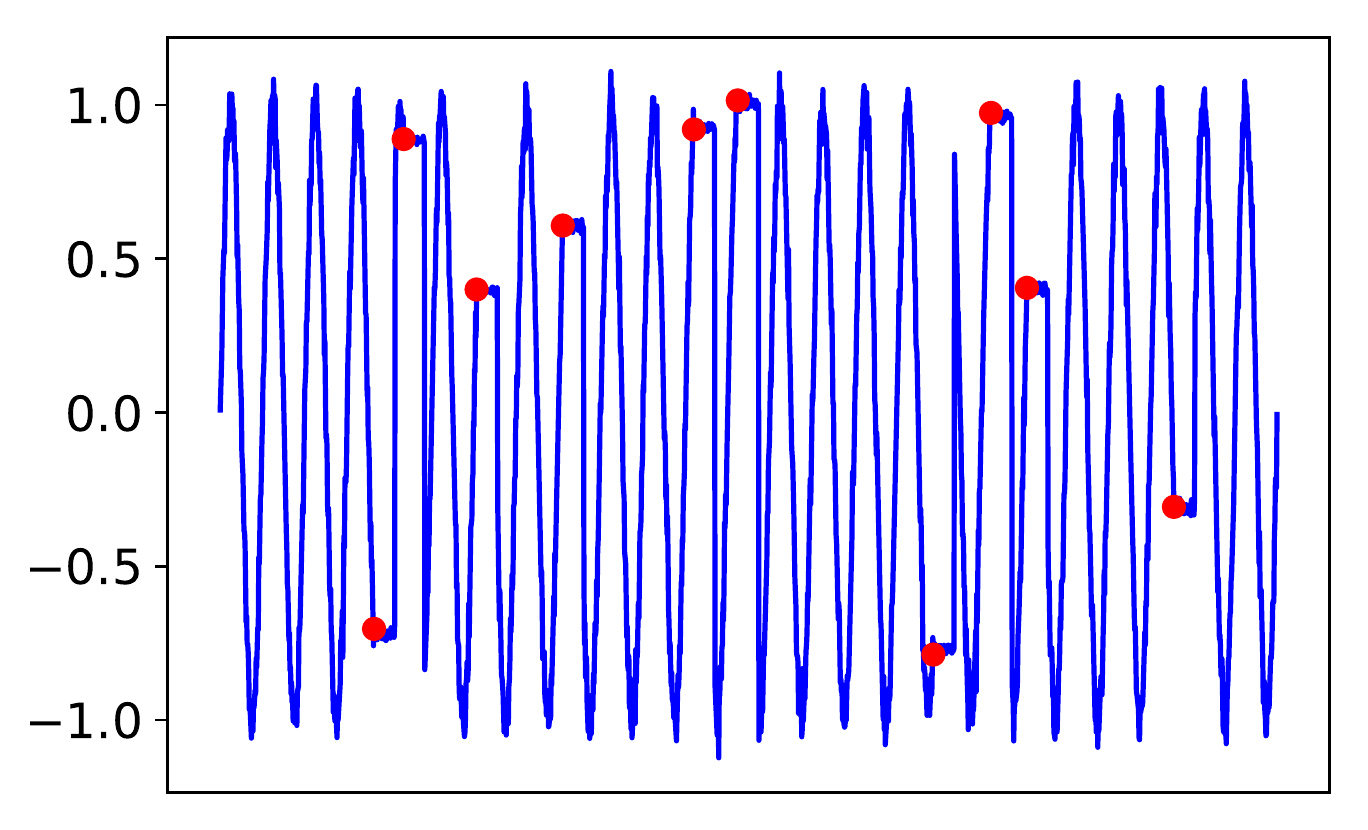}
		\caption{An example of a time series, the red dots represent the beginning of an anomalous segment.}
		\label{fig:ts-example}
	\end{subfigure}	\hfill
	\begin{subfigure}{0.45\textwidth}
		\centering
		\includegraphics[width=\textwidth]{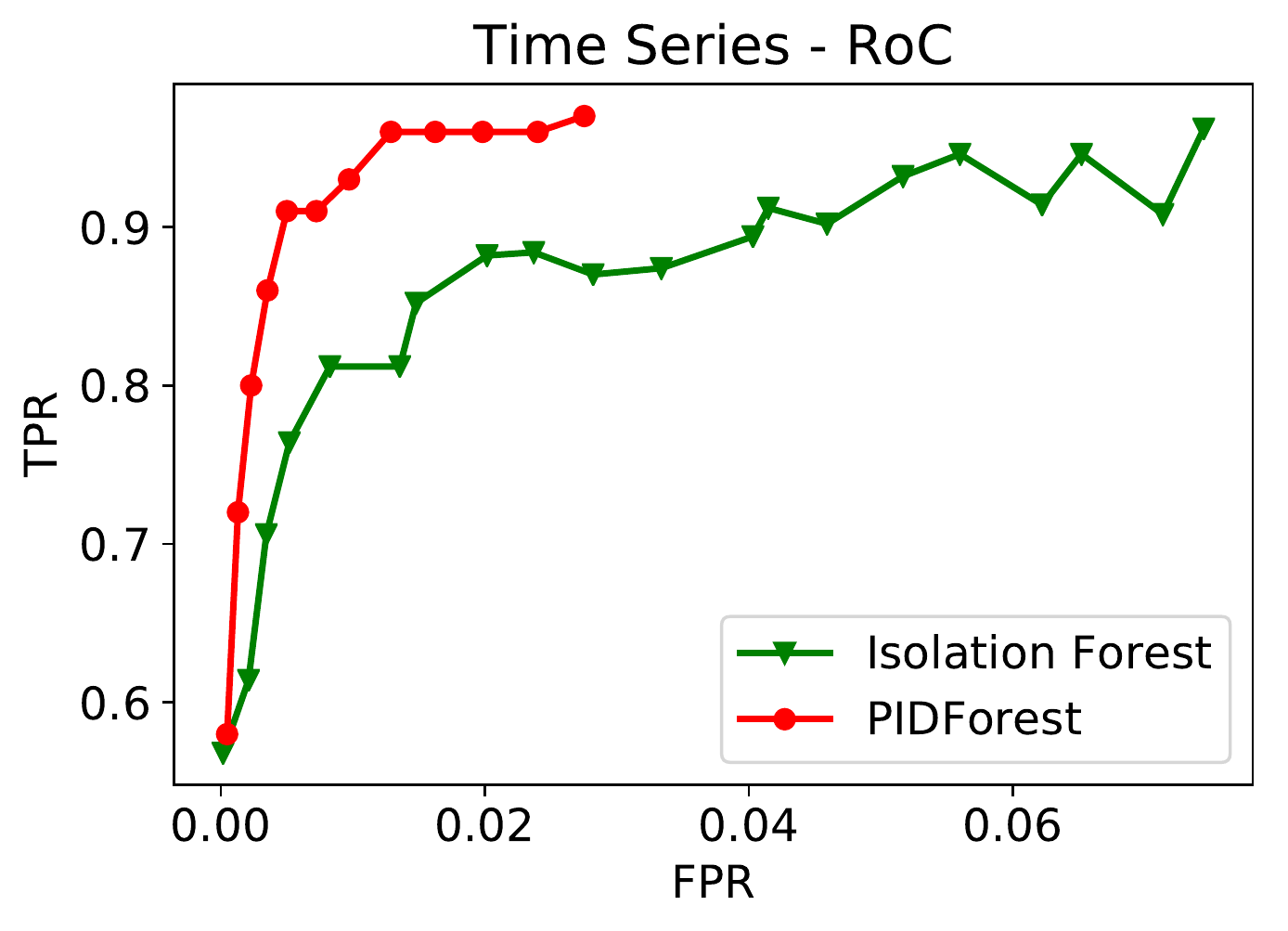}
		\caption{Performance comparison on time series data.}
		\label{fig:ts_exp}
	\end{subfigure}
	\caption{Synthetic experiments on time series data.}
\end{figure*}

We create a periodic time series using a simple
$\sin$ function with a period of $40$. We choose $10$ locations and
fix the value for the next $20$ points following each of these
locations. These regions are the anomalies. Finally we add small
Gaussian noise to the series. See Fig.~\ref{fig:ts-example} for an example. As in
Section \ref{sec:real}, we shingle the time series with a window of 10.
Fig.~\ref{fig:ts_exp} shows the ROC curve (true positive rate (TPR)
vs. false positive rate (FPR)), averaged over $10$ runs. Since all
dimensions are a priori identical, choosing splits at random seems
natural. So we expect \iF\ to perform  well, and indeed it achieves a precision of almost
$1$ while catching $5$ out of the $10$ anomalies. However, \iF\ struggles to catch all anomalies, and \density has a
significantly better precision for high recall.

%We then calculated the true positive rate (TPR) and false positive rate (FPR), where we deemed a point a true positive if a shingle it belongs to intersects an anomalous region. The results can be seen in

\subsection{Robustness to Choice of Hyperparameters}\label{sec:hyper}

\begin{figure}
	\begin{subfigure}{0.45\textwidth}
		\centering
		\includegraphics[width=\textwidth]{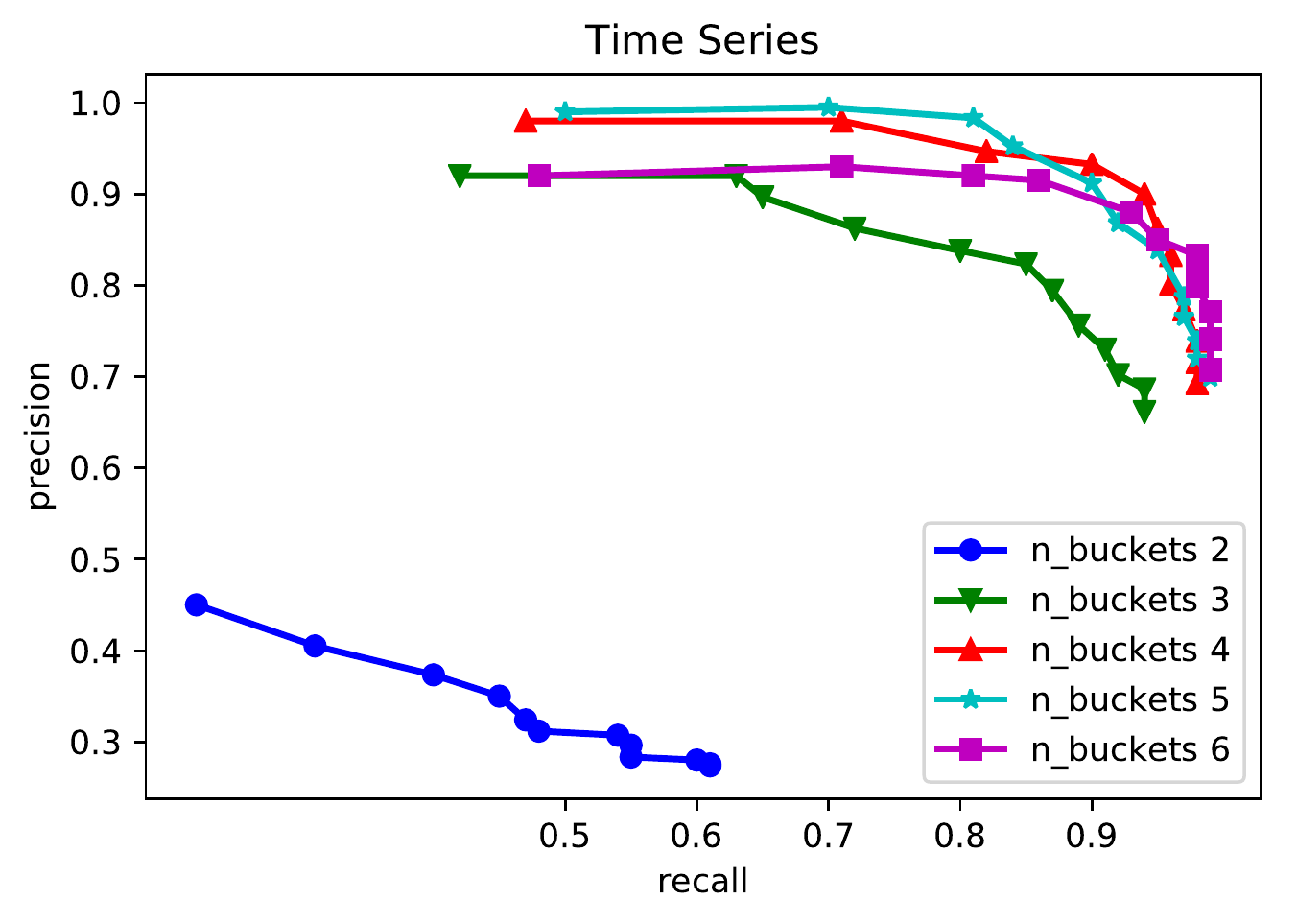}
		\caption{Varying number of buckets}
		\label{fig:hyper-1}
	\end{subfigure}\hfill
	\begin{subfigure}{0.45\textwidth}
		\centering
		\includegraphics[width=\textwidth]{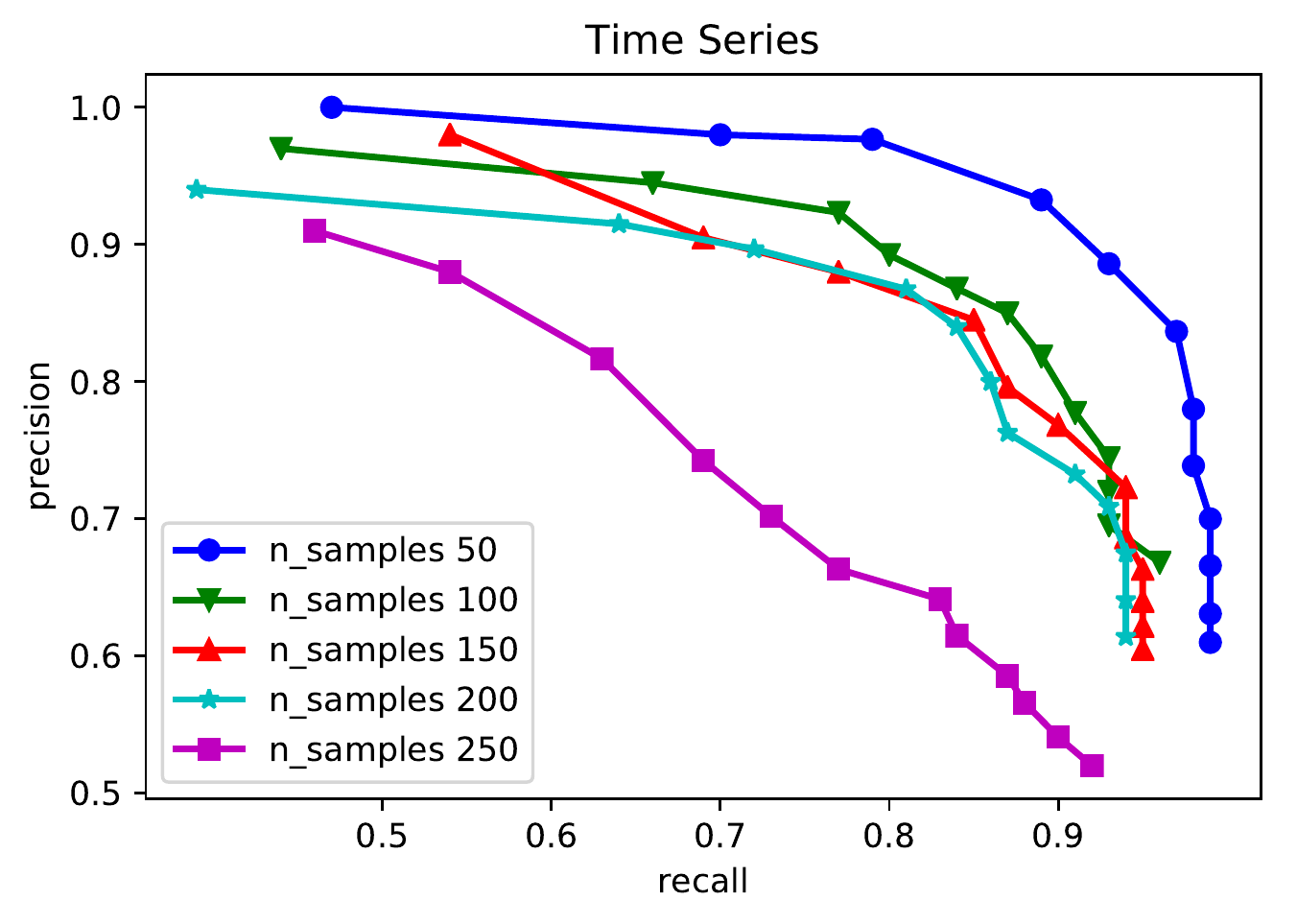}
		\caption{Varying number of samples}
		\label{fig:hyper-2}
	\end{subfigure}
	\label{fig:hyper}
	\caption{Robustness to choice of hyperparameters}
\end{figure}

One of the appealing properties of PIDForest is that the quality of
the output is relatively insensitive to the exact value of the
hyper-parameters. We tested multiple settings and found that
each hyper-parameter has a moderate value above which the quality of
the output remains stable. Figure~\ref{fig:hyper-1} shows precision-recall in the synthetic
time-series experiment, where we vary the parameter $k$ (number of
buckets). We see that $k = 2$ is too little, but there is relatively small variation between $ k = 3,4,5,6$. Similar behavior was observed for
the number of samples $m$  (see Figure~\ref{fig:hyper-2}), number of trees $t$ and depth of each tree $h$, and
also with the mixture of Gaussians experiment. Since these parameters
do affect the running time directly, we set them to the smallest
values for which we got good results.

\section{Relation to prior work}
\label{sec:related}

Anomaly detection is a widely studied area with a number of comprehensive surveys and books
\cite{ChandolaBK09,HodgeA04,Patcha2007}. In the
following we discuss the basic techniques and how they relate to our
work.

\textbf{Proximity based methods:}
Many works detect anomalies by computing distances to nearest
neighbors in various methods \cite{BandaragodaTALZ18,Breunig:2000,Angiulli:2002:FOD:645806.670167,Ramaswamy:2000:EAM:335191.335437,Shi2006}. There are many different methods with pros and cons,
which we do not discuss here. The important common feature is that
these algorithms utilize a notion of \emph{distance}, typically
Euclidean, as a proxy for similarly across points. Once the notion of
distance is established it is used find the closest neighbor, or the
close $k$ neighbors, it is used to define a ball of a given radius
around a point and so on.

There are few major drawbacks with these approaches. First,  as the
number of dimensions increases the distances between points become
more and more similar, and notion of a local neighborhood looses its
meaning. Further, distance is an ill suited notion to the general case
where some columns are categorical and some numeric and where we
different columns employ different units. Isolation based algorithms
work with the input basis, they do not compute linear combinations of
attributes which are required to change basis. This is an advantage in
dealing with heterogeneous data.

Having said that, for homogeneous datasets that involve audio or visual inputs (like Vowels/MNIST),
the input basis may not be the {\em right} basis. If the signal has
sparsity in some special basis, then $\ell_2$ distance, or some
variant of it like Mahalanobis distance might be a natural metric for
such settings. In such situations kNN, PCA might do better than
isolation based approaches.

\textbf{Density based algorithms:}
Density has been used as a criterion for several works on
clustering. For instance, DBSCAN \cite{Ester:1996} builds clusters from maximal connected
components out of dense balls (in a chosen metric) of a certain
radius. Outliers are points that do not belong in any such
cluster.  The work of Agrawal {\em et al.} \cite{AgrawalGGR98} builds
clusters that can be expressed as connected unions of subcubes.
A key difference from these works is that we do not
attempt to discover the structure (such as clusters) in the normal
data. Further, rather than only consider balls/subcubes at a
particular scale (this scale is a hyperparameter), our algorithms
attempt to minimize density over subcubes at all scales.

\textbf{Isolation Forest: } The most relevant algorithm is the widely used Isolation
Forest \cite{iForest}, a detailed comparison of the technical
differences between the two algorithms is
in Section  \ref{sec:if}.
Since \cite{iForest} chooses \emph{randomly} which column to split,
this causes Isolation Forest's accuracy to degrade when
extra dimensions are introduced (this is a well known issue, see
\cite{GuhaMRS16,BandaragodaTALZ18}). Robust Isolation Forest
\cite{GuhaMRS16} deals with this by choosing a column based on the
size of its range. This makes the algorithm scale-sensitive and
results change based on the units with which the data is reported.
\density on the other hand zooms on the columns which have the most
\emph{signal}, which makes it robust to noisy and irrelevant dimensions. Another
difference is that Isolation Forest insists on full isolation and then
computes the average leaf depth, which makes it sensitive to
duplicates or near duplicates. See  Section~\ref{sec:synthetic} where
we demonstrate these points via experiments.

\eat{

The set of encodings or descriptions we allow could be more complex
(say a linear separator or a neural net). We make a conscious choice
above to work with conjunctions of attributes. There are multiple
reasons for this design choice:
\begin{itemize}
   \item {\em Conjunctions are explainable.} Explanations in terms of
    conjunctions of attribute values as described above seem to best
    conform to our intuitive notion for what is anomalous. Using a
    richer hypothesis class might make it harder to explain why a
    particular example is labeled anomalous.

  \item {\em Sometimes less is more. } Allowing a more expressive
     set of descriptions would mean shorter descriptions for every
     entry in our table. This could have the undesired effect of
   reducing the range of description lengths, making the task of
   separating anomalous points harder.

 \item {\em Algorithmic efficiency. } As we will see, the proposed
     notion lends itself to a highly efficient algorithm based on
     decision forests.
\end{itemize}
}

%\vspace{-8pt}
\section{Conclusion }

We believe that \pidF\ is arguably one of the
best off-the-shelf algorithms for anomaly detection on a
large, heterogenous dataset. It inherits many of the desirable
features of Isolation Forests, while also improving on it in
important ways. Developing provable and scalable approximations to
$\pidsc$ is an interesting algorithmic challenge.

%\section*{Acknowledgment}
%We thank Ananya Kumar, Daniel Kang, Rob Johnson and the anonymous reviewers for detailed feedback on initial versions of the paper. 

\bibliographystyle{unsrtnat}
\bibliography{anomaly}

\appendix
\section{Finding optimal splits efficiently}
\label{sec:splitting}

In this section we present the algorithm used to split a single
dimension. The problem is easy in the discrete setting, so we focus on the continuous case.
We first restate the problem we which solve. In what follows, we
may assume the dataset $\P \subseteq I$ is one-dimensional, since we
work with the projection of the dataset onto the dimension $j$.

For interval $I$, a $k$-interval partition of $I$ is a partition into
a set $\{I_1, \ldots, I_k\}$ of disjoint intervals. Letting
$q_i \defeq |P \cap I^i|/|P|$, our goal is to maximize
\begin{align}
\label{eq:var-pk2}
  \Var(\P,\{I^i\}_{i \in [k]}) = \sum_{i \in [k]}q_i\rho(I^i)^2 - \rho(I)^2.
\end{align}
Since the term $\rho(I)^2$ is independent of the partition, we can
drop it and restate the problem as follows.

{\bf Optimal $k$-split: }
Given a set $\P$ of $m$ points $x_1 \leq  \cdots \leq  x_m$
from an interval $I$, and a parameter $k$, find a $k$-interval partition of $I$
so as to maximize
\begin{align}
  \label{eq:cost-pk}
  \cost(\P, k) =  \sum_{i=1}^kq_i\rho(I_i)^2.
\end{align}

By shifting and scaling, we will assume that the bounding interval $I = \intzo$.
We also assume the $x_i$s are distinct (this is not needed, but eases notation).
To simplify matters, we restrict to those intervals whose start and end points
are either $e_0 = 0$, $e_m =1$, or $e_i = (x_i +x_{i+1})/2$ for $i \in
[m-1]$.  This avoids issues that might arise from the precision of the end points,
and from having points lie on the boundary of two intervals (to which
interval should they belong?). It reduces the search space of
intervals to $O(m^2)$. One can use dynamic programming to give an
$O(m^2k)$ time and $O(mk)$ space algorithm to solve the problem
exactly. However this is too costly, since the procedure runs in an
inner loop of \pidF\ Fit. Rather we show the problem reduces to
that of computing optimal $k$-histograms for an array, for which
there are efficient streaming approximation algorithms known.

First some notation. An interval in $[m]$ is  $J = \{i: \ell \leq i \leq
u\}$, and $|J| = u - \ell + 1$. Given a function $f:[m] \rgta
\R$ and an interval $J \subseteq [m]$, let $\ol{f}(J) = \sum_{i \in
  J}f(i)/|J|$ denote the average of $f$ over the interval $J$. A
$k$-interval partition of $[m]$ is a set of pairwise disjoint
intervals $\{J_1, \ldots, J_k\}$ whose union is $[m]$. Given  $j \in
[m]$, let $J(j) \in \I$ denote the interval containing it.

{\bf Optimal $k$-histograms:}
Given $f:[m] \rgta \R$, find a $k$-interval partition of $[m]$
which maximizes
\begin{align}
  \label{eq:cost-fk}
  \cost(f, k) = \sum_{i \in [k]}|J_i|(\ol{f}(J_i))^2.
\end{align}
We show in Lemma \ref{lem:eq} that this is equivalent to finding the
$k$-histogram that minimizes the $\ell_2$-error.
We now give the reduction from computing $k$-splits to
$k$-histograms. Recall that $e_0 = 0$, $e_m =1$, or $e_i = (x_i +x_{i+1})/2$ for $i \in
[m-1]$.  For each $i \in [m]$, let $f(i) = e_i - e_{i-1}$
denote the length of the interval $[e_{i-1}, e_i]$ which contains
$x_i$. There is now a natural correspondence between the discrete interval
$J_{\ell, u} =\{\ell, \cdots, u\}$ and the continuous interval
$I_{\ell, u} = [e_{\ell-1}, e_u]$ which contains the points $\{x_\ell,
\cdots, x_u\}$ from $\P$, where
\[ \ol{f}(J_{\ell, u}) = \frac{\sum_{i=\ell}^u (e_i - e_{i-1})}{u - \ell + 1} =
\frac{e_u - e_{\ell -1}}{u -\ell +1} = \rho(I_{\ell, u}) \]
Thus a $k$ interval partition of $[m]$ translates to a $k$-interval
partition of $I$, with objective function
\[ \cost(f, k) = \sum_{i \in [k]}|J_i|(\ol{f}(J_i))^2 = \sum_{i \in
  [k]}|\P \cap I_i|\rho(I_i)^2 = m\cdot \cost(\P, k). \]

An efficient streaming algorithm for computing approximately optimal
$k$-histograms is given by Guha {\em et al.} \cite{GuhaKS06}, which
requires space $O(m + k^2)$ and time $O(m + k^3)$ (we set their
parameter $\eps$ to $0.1$). We use their algorithm in the Fit procedure to
find good splits.

\paragraph{Comparison with exact dynamic programming algorithm.} As discussed, the variance maximization problem also admits a much less
efficient but exact dynamic programming algorithm. On a 300
dimensional array, the algorithm of \cite{GuhaKS06} is 50x faster than
the exact dynamic program in finding a best 5-bucket histogram (0.05 s
vs 2.5s). This is as expected given that the complexity of the DP
is $O(m^2)$, whereas the approximate algorithm takes time
$O(m)$. %where $m$ is the number of samples.
\section{Proofs}
\label{app:proof}

\begin{Lem}
  \label{lem:dense-01}
  For $x \in \T$,
  \[ \max_{C \ni x}\rho_{0,1}(\T, C) = 2^{d - \pid(x, \T)}.\]
\end{Lem}
\begin{Proof}
Given $S \subseteq [d]$, let
\[ C_x(S) = \{ y \in \zo^d  \ s.t. \ y_S = x_S \} \]
be the subcube consisting of $2^{d - |S|}$ points that agree with $x$ on
$S$.
Since $C_x(S) \cap \T = \Imp(x, \T, S)$,
\[ \rho_{0,1}(\T, C_x(S)) = \frac{|C_x(S)|/|C_x(S) \cap \T|}{|\T|} =
\frac{2^{d - |S|}}{|\Imp(x, \T, S)|} = 2^{d - (|S| + \log_2(|\Imp(x, \T,  S)|))}\]
Iterating over all subsets $S$ gives all the subcubes that contain
$x$. The RHS is minimized by taking $S = \PID(x, \T)$ by Definition
\ref{def:boolean-pid}. This gives the desired result.
\end{Proof}

\begin{Lem}
  \label{lem:eq}
  The $k$-histogram which maximizes the cost minimizes the squared $\ell_2$error.
\end{Lem}
\begin{Proof}
Given a $k$-interval partition $\{J_1, \ldots, J_k\}$, the minimum
$\ell_2$ error is obtained by approximating $f$ by its average over
each interval. The squared $\ell_2$ errors is given by
\begin{align*}
  \sum_{j \in m}(\ol{f}(J(j)) - f(j))^2 &=
\sum_{j \in [m]} (f(j)^2 - 2\ol{f}(J(j))f(j) + \ol{f}(J(j))^2)\\
&= \sum_{j \in [m]} f(j)^2 - \sum_{i \in [k]}2\ol{f}(J_i)\sum_{j \in J_i}f(j) + \sum_{i \in [k]}\sum_{j \in J_i}\ol{f}(J(j))^2)\\
&= \sum_{j \in [m]}f(j)^2 -  2\sum_{i \in [k]}|J_i|(\ol{f}(J_i))^2 +
\sum_{i \in [k]}|J_i|(\ol{f}(J_i))^2\ \ \ \text{since} \ \sum_{j \in J_i}f(j) = |J_i|\ol{f}(J_i)\\
& = \sum_{j \in J}f(j)^2 - \sum_{i \in [k]}|J_i|(\ol{f}(J_i))^2\\
& = \sum_{j \in J}f(j)^2 - \cost(f, k)
\end{align*}
Hence  maximizing $\cost(f, k)$ is equivalent to minimizing
the squared error of the histogram.
\end{Proof}

%\section{Experiments on real-world datasets} \label{app:real}

\begin{figure}

	\begin{subfigure}{0.45\textwidth}
		\centering
		\includegraphics[width=\textwidth]{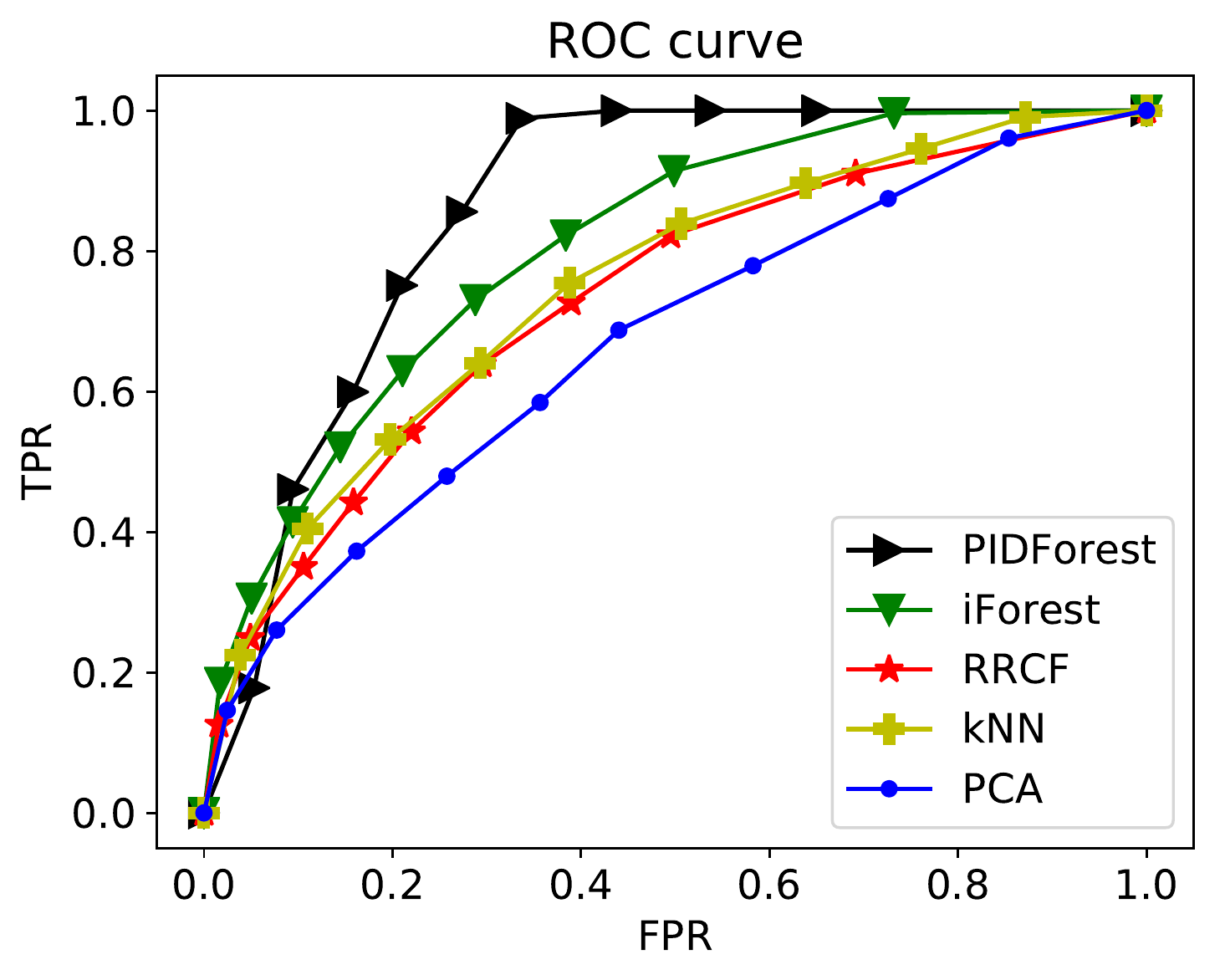}
		\caption{ROC curve for \emph{Thyroid} dataset.}
		\label{fig:thyroid}
	\end{subfigure}\hfill
	\begin{subfigure}{0.45\textwidth}
		\centering
		\includegraphics[width=\textwidth]{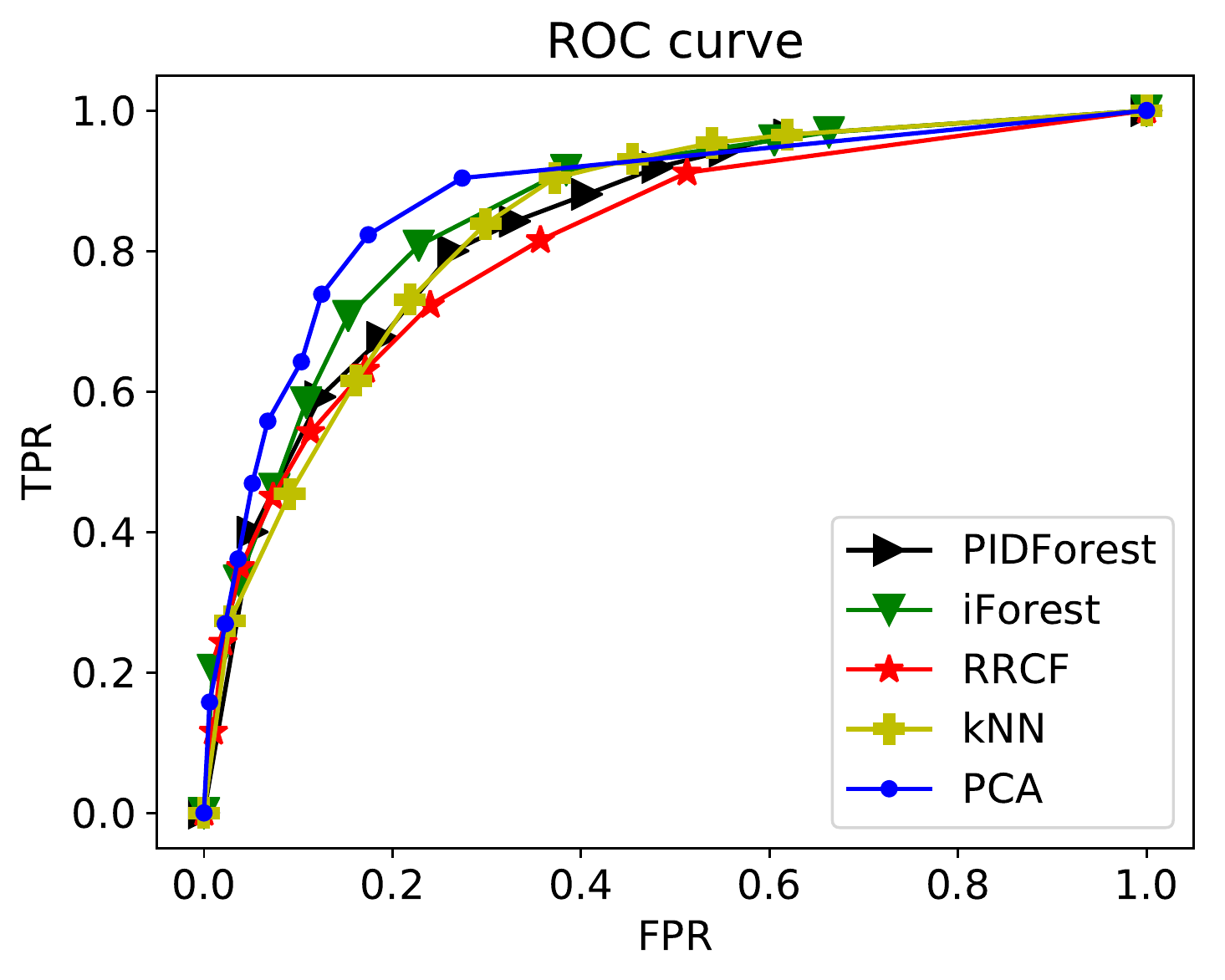}
		\caption{ROC curve for \emph{Mammography} dataset.}
		\label{fig:mammography}
	\end{subfigure}

		\begin{subfigure}{0.45\textwidth}
			\centering
			\includegraphics[width=\textwidth]{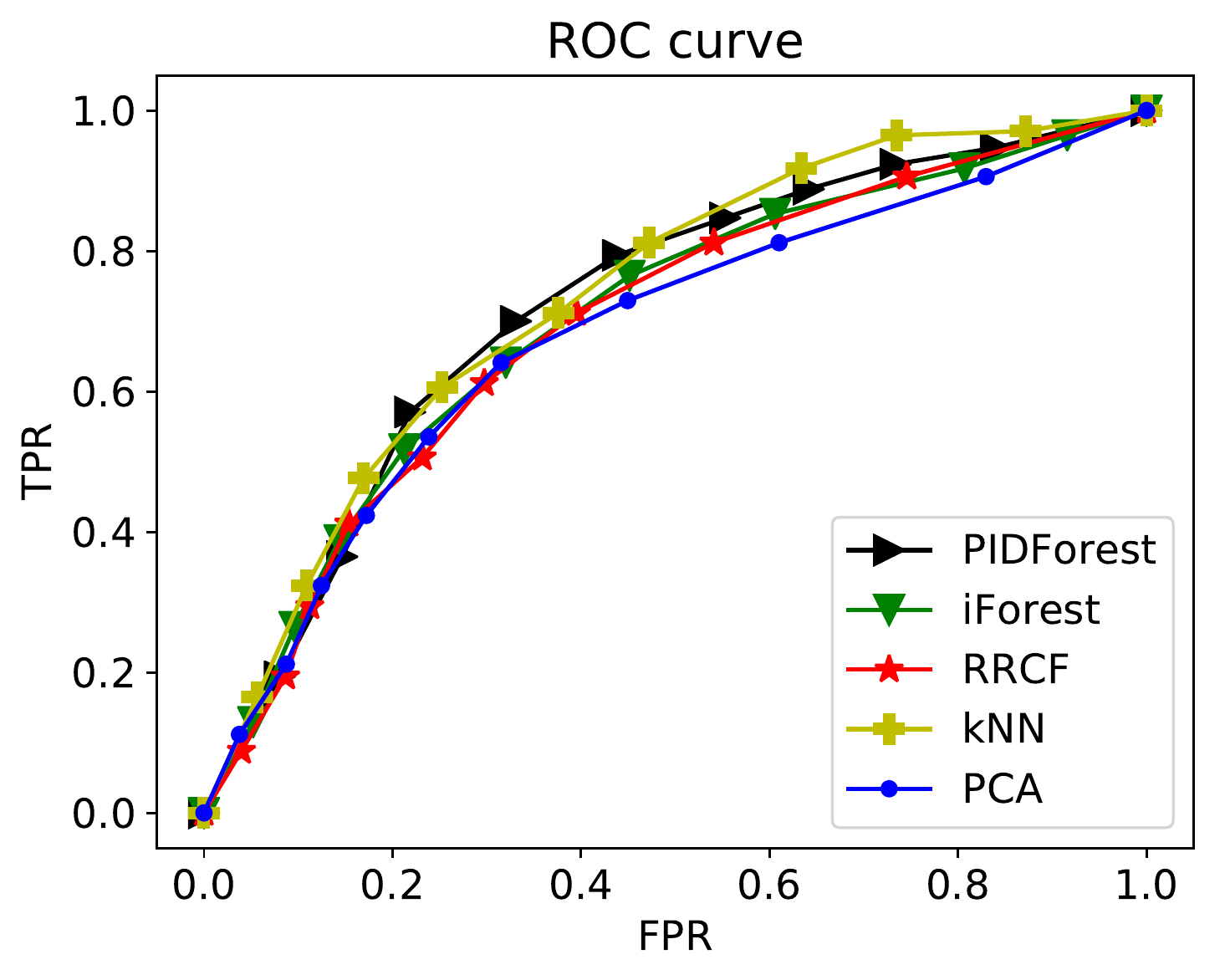}
			\caption{ROC curve for \emph{Siesmic} dataset.}
			\label{fig:siesmic}
		\end{subfigure}\hfill
		\begin{subfigure}{0.45\textwidth}
			\centering
			\includegraphics[width=\textwidth]{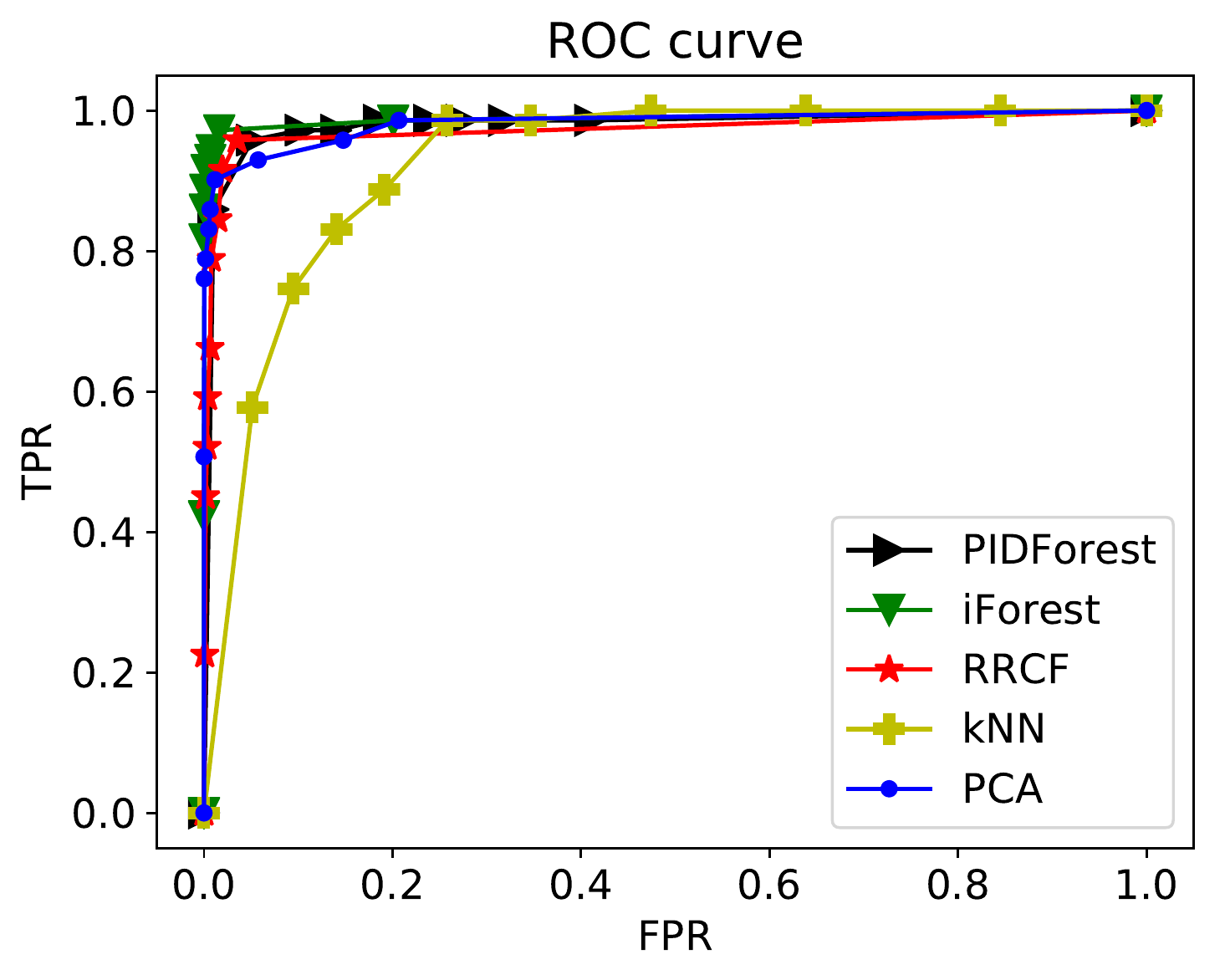}
			\caption{ROC curve for \emph{Satimage-2} dataset.}
			\label{fig:satimage}
		\end{subfigure}

			\begin{subfigure}{0.45\textwidth}
				\centering
				\includegraphics[width=\textwidth]{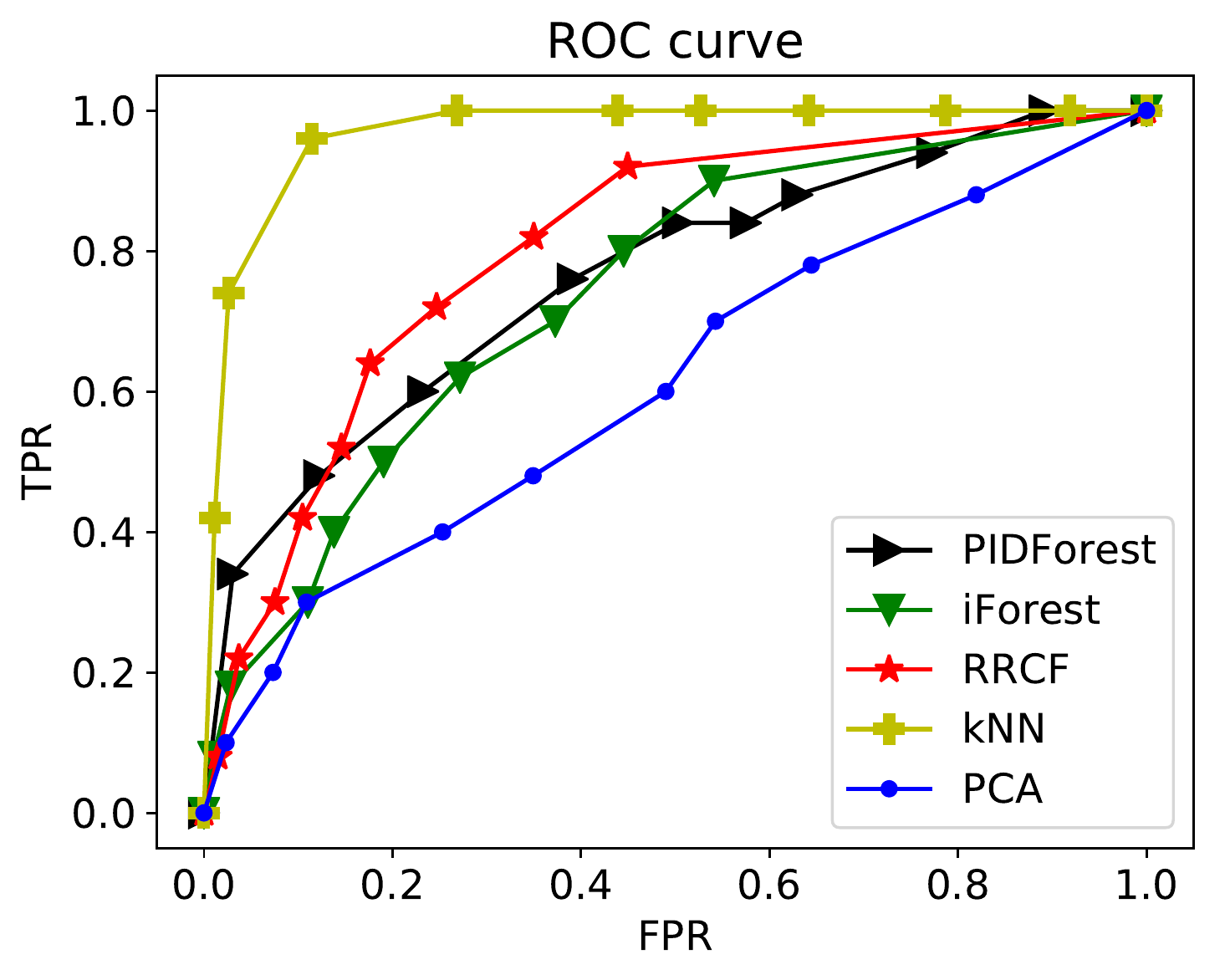}
				\caption{ROC curve for \emph{Vowels} dataset.}
				\label{fig:vowels}
			\end{subfigure}\hfill
			\begin{subfigure}{0.45\textwidth}
				\centering
				\includegraphics[width=\textwidth]{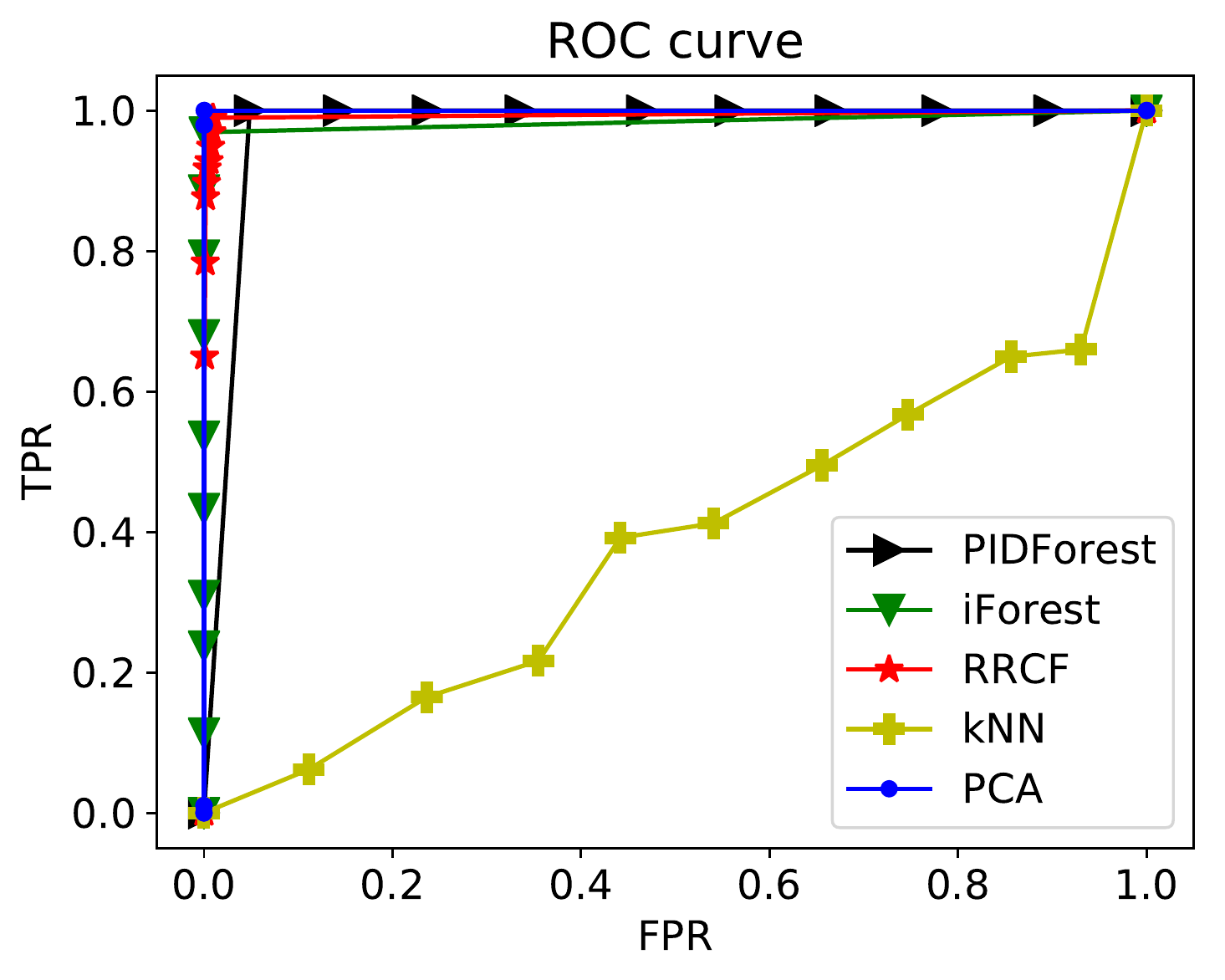}
				\caption{ROC curve for \emph{Musk} dataset.}
				\label{fig:musk}
			\end{subfigure}

	\caption{ROC curves for the first six datasets from Table \ref{tab:results}. For visual clarity, we omit LOF and SVM which did not perform as well as the other algorithms.}
	\label{fig:roc1}
\end{figure}

\begin{figure}

	\begin{subfigure}{0.45\textwidth}
		\centering
		\includegraphics[width=\textwidth]{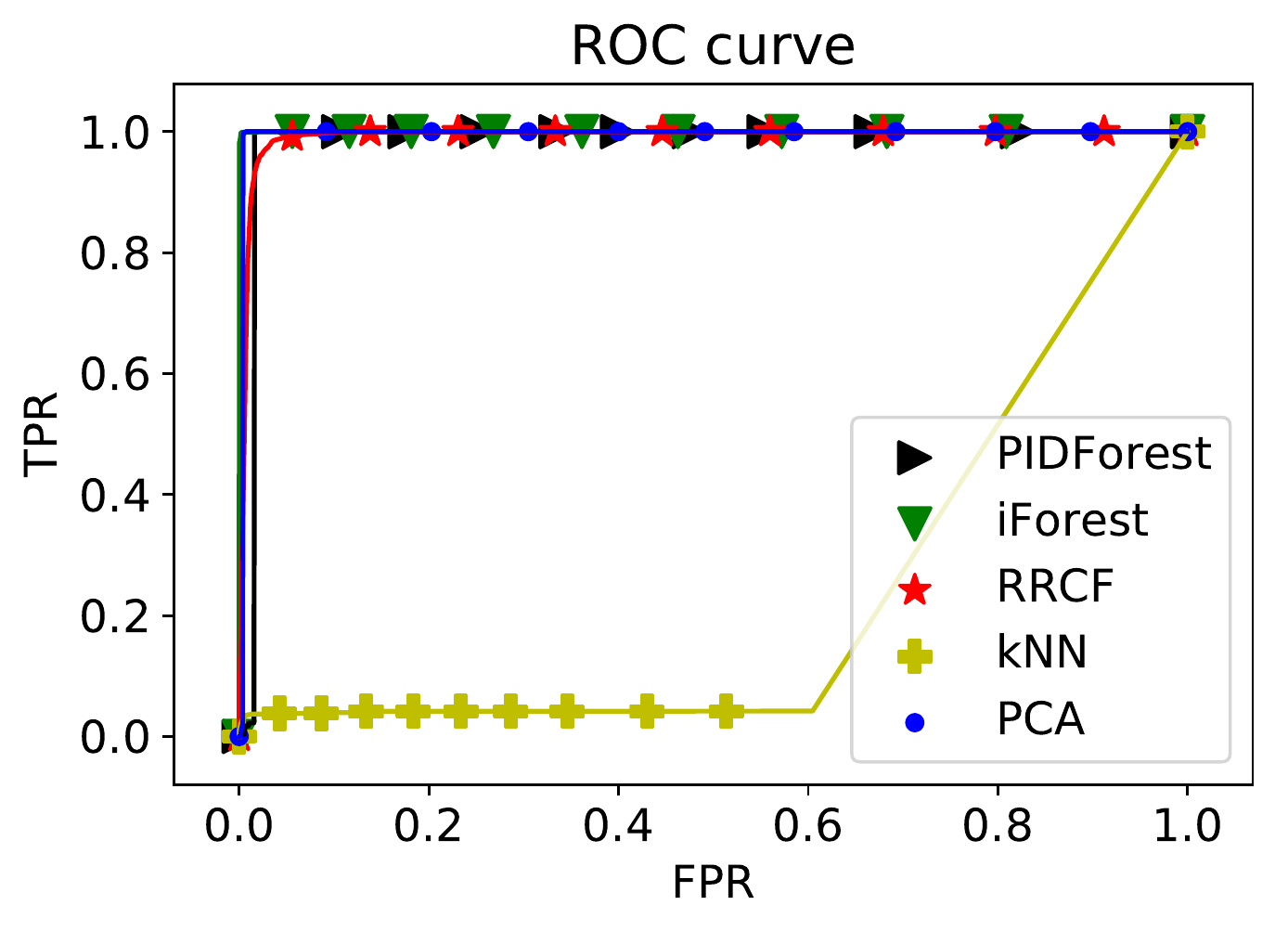}
		\caption{ROC curve for \emph{http} dataset.}
		\label{fig:http}
	\end{subfigure}\hfill
	\begin{subfigure}{0.45\textwidth}
		\centering
		\includegraphics[width=\textwidth]{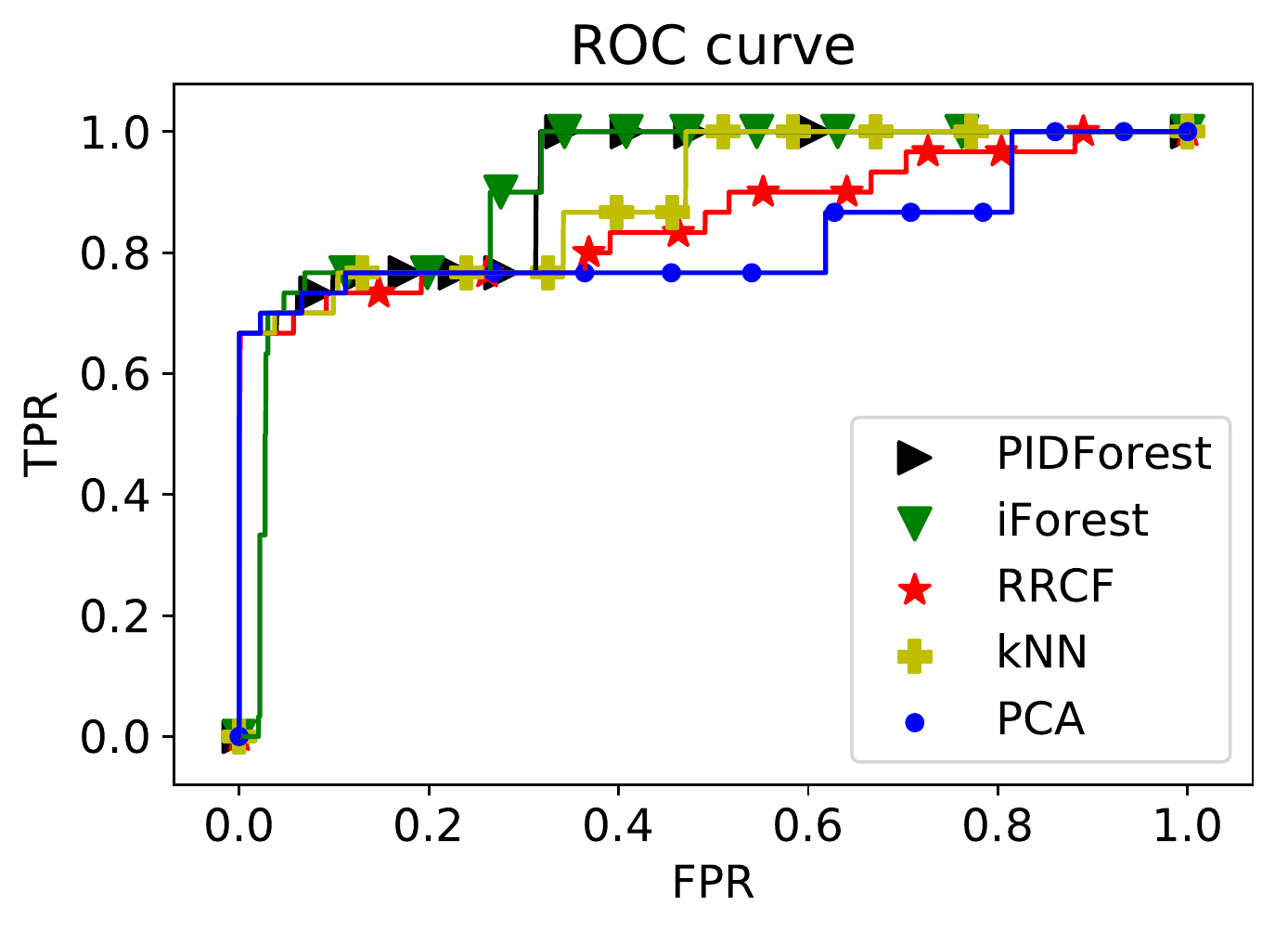}
		\caption{ROC curve for \emph{smtp} dataset.}
		\label{fig:smtp}
	\end{subfigure}

	\begin{subfigure}{0.45\textwidth}
		\centering
		\includegraphics[width=\textwidth]{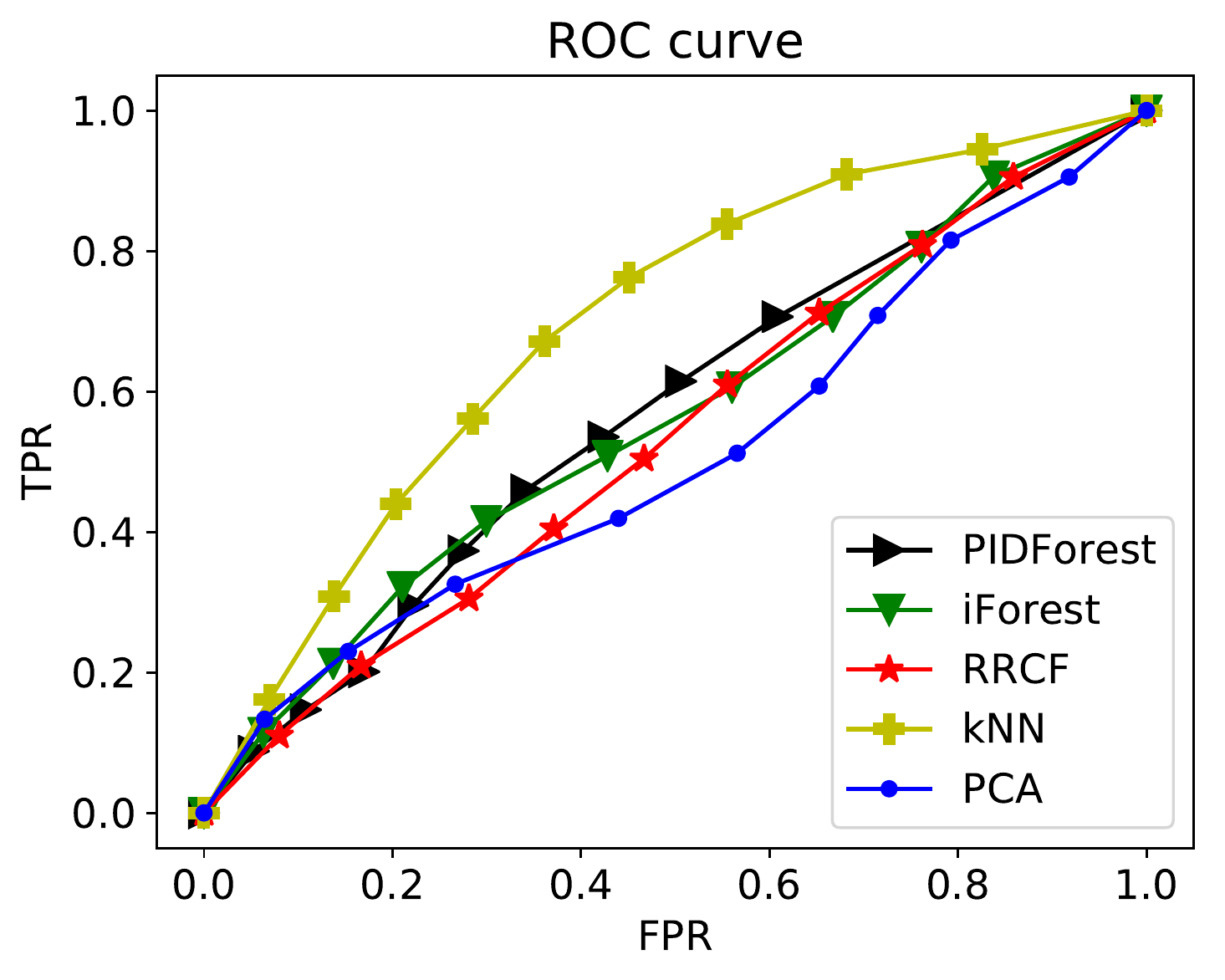}
		\caption{ROC curve for \emph{NYC taxi} dataset.}
		\label{fig:nyc}
	\end{subfigure}\hfill
	\begin{subfigure}{0.45\textwidth}
		\centering
		\includegraphics[width=\textwidth]{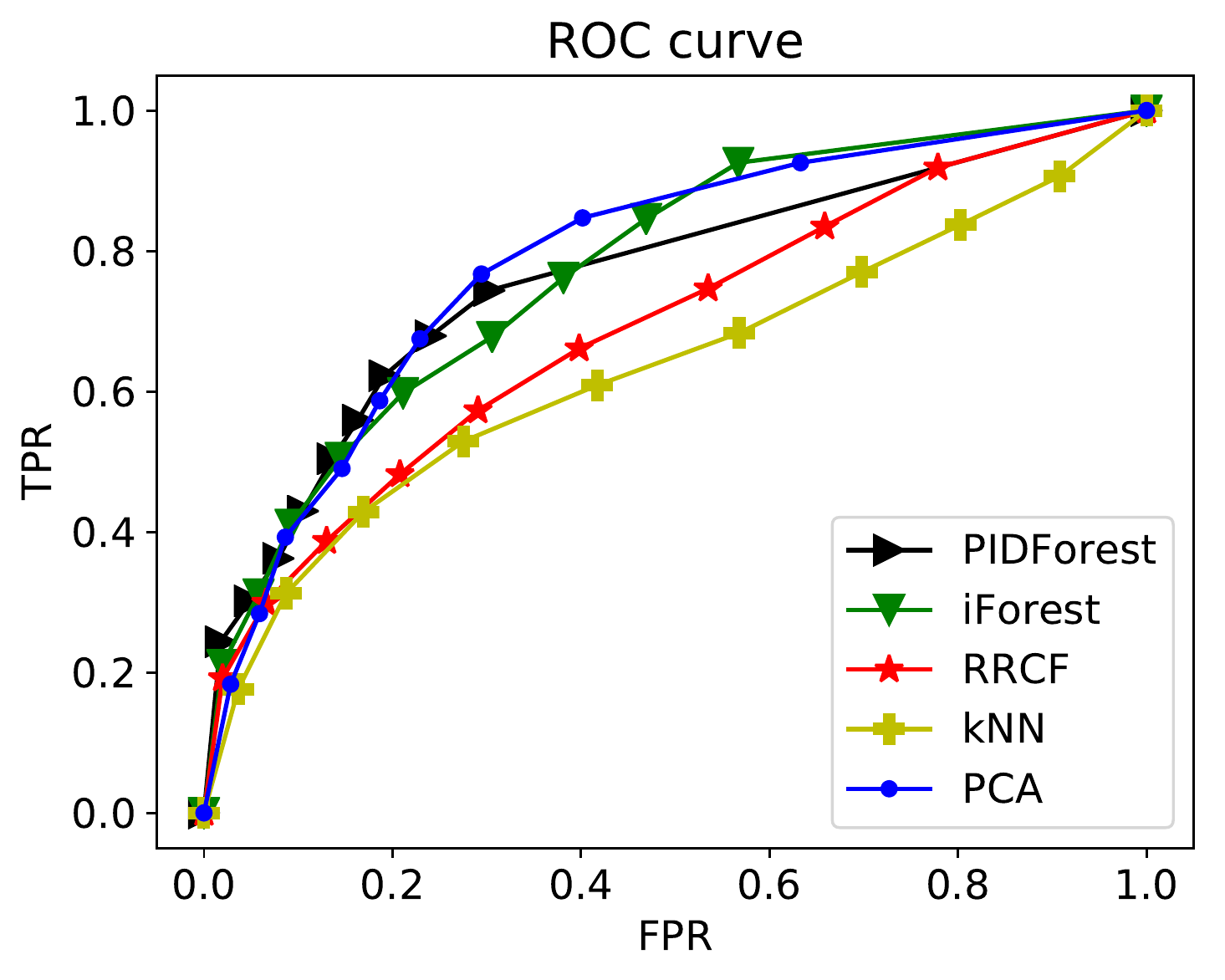}
		\caption{ROC curve for \emph{Ambient temperature} dataset.}
		\label{fig:ambient}
	\end{subfigure}

	\begin{subfigure}{0.45\textwidth}
		\centering
		\includegraphics[width=\textwidth]{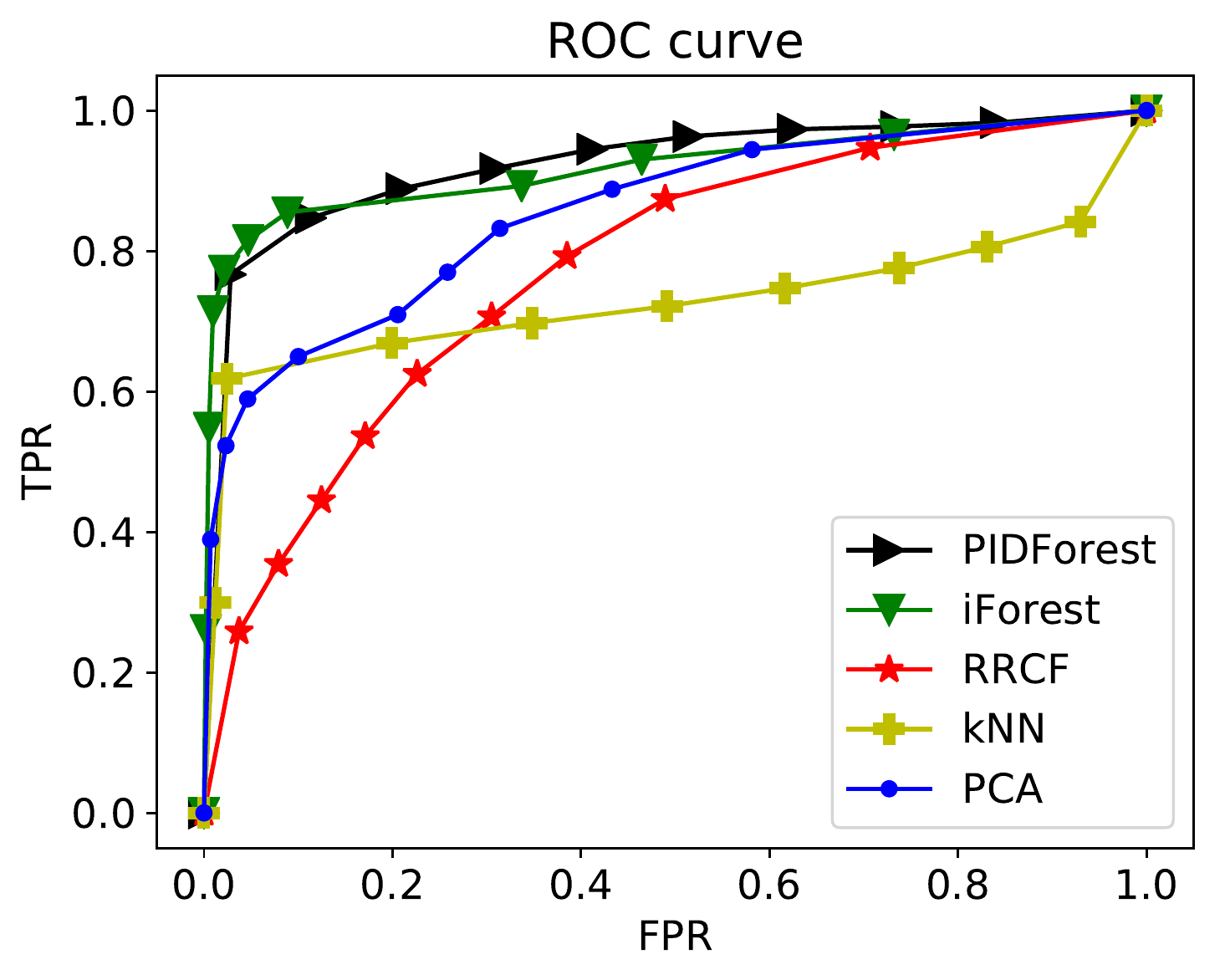}
	\caption{ROC curve for \emph{CPU utilization} dataset.}
	\label{fig:cpu}
	\end{subfigure}\hfill
	\begin{subfigure}{0.45\textwidth}
		\centering
		\includegraphics[width=\textwidth]{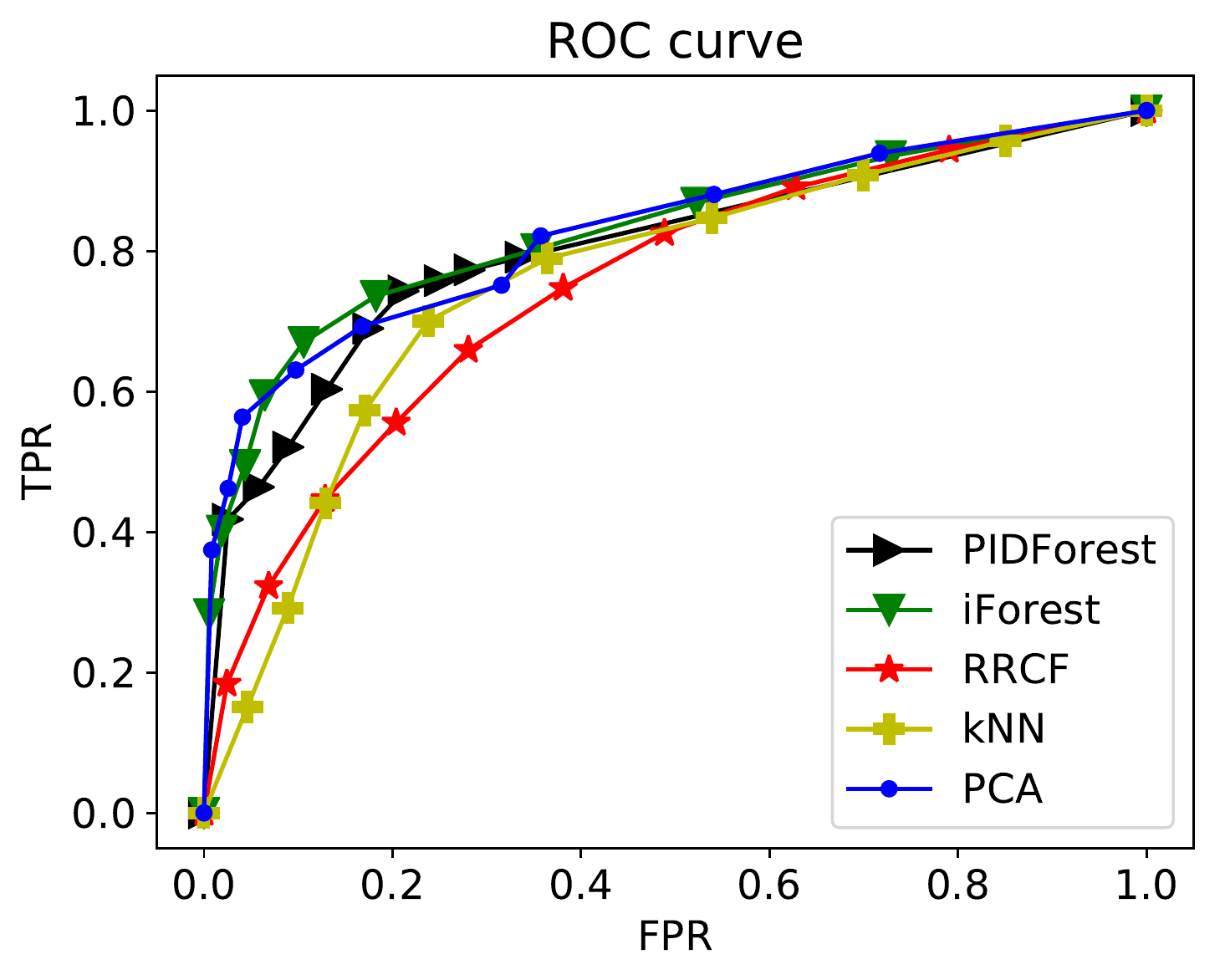}
		\caption{ROC curve for \emph{Machine temperature} dataset.}
		\label{fig:machine}
	\end{subfigure}

	\caption{ROC curves for the last six datasets from Table \ref{tab:results}. For visual clarity, we omit LOF and SVM which did not perform as well as the other algorithms.}
	\label{fig:roc2}
\end{figure}

\end{document}